\pdfoutput=1
\documentclass[11pt]{article}

\usepackage[preprint]{coling}

\usepackage{times}
\usepackage{latexsym}

\usepackage[T1]{fontenc}
\usepackage[utf8]{inputenc}

\usepackage{microtype}
\usepackage{amsmath}
\usepackage{amsfonts}
\usepackage{graphicx}
\usepackage{inconsolata}

\usepackage{geometry}
\usepackage{tabularx}
\usepackage{xtab}
\usepackage{makecell}
\title{Steering Conversational Large Language Models for Long Emotional
Support Conversations}

\author{Navid Madani \and Sougata Saha \and Rohini Srihari \\
        Computer Science and Engineering - University at Buffalo \\
        Buffalo, NY, 14260 \\
        \texttt{\{smadani, sougatas, rohini\}@buffalo.edu} }

\begin{document}
\maketitle
\begin{abstract}
In this study, we address the challenge of enabling large language models (LLMs) to consistently adhere to emotional support strategies in extended conversations. We focus on the steerability of the Llama-2 and Llama-3 suite of models, examining their ability to maintain these strategies throughout interactions. To assess this, we introduce the Strategy Relevant Attention (SRA) metric, which quantifies the model's adherence to the prompted strategy through attention maps. To facilitate our study, we create a strategy-conditioned synthetic conversational dataset derived from the ESConv dataset. We also propose various baselines informed by our proposed SRA metric to address the challenge and propose a fine-tuned model that significantly enhances the steerability of the base model in following the strategy throughout the conversation. The code and data are publicly available on our GitHub. \footnote{
\href{https://github.com/navidmdn/ESConv-SRA}{https://github.com/navidmdn/ESConv-SRA}}

\end{abstract}

\begin{figure}[!ht]
    \centering
    \includegraphics[width=0.47\textwidth]{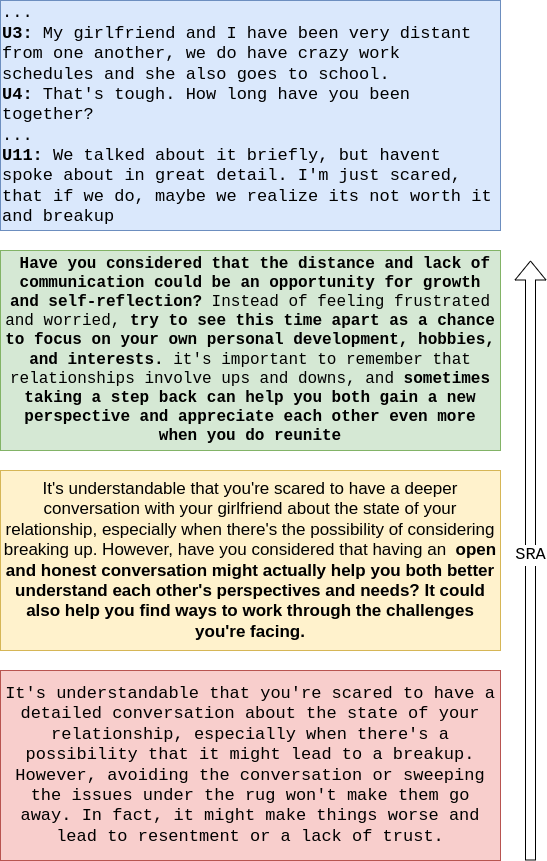} % Adjust the path to your image file and the scaling as needed
    
    \caption{A sample continuation of a conversation using \textbf{"Provide Different Perspectives"} strategy, given by three different prompt templates sorted by the SRA metric increasing from bottom to top using Llama-70b-chat model. The model output using the prompt template with higher SRA adheres better to the given strategy.}
    \label{fig:fig1} % Label for referencing the figure in text
\end{figure}

\section{Introduction}
In the rapidly evolving domain of conversational AI, the creation of emotionally intelligent conversational agents is becoming increasingly important as it opens up new possibilities for more natural and helpful interactions between humans and machines. Central to this transformative journey is the challenge of empowering large language models (LLMs) not only to partake in natural dialogues but also to adeptly navigate and influence the conversation flow using expert strategies derived from psychology and emotional support literature.

 This paper delves into the critical aspect of integrating emotional support strategies into conversational LLMs, a domain that remains largely uncharted yet holds significant promise for a range of applications, from mental health support to customer service.

The advent of the Emotional Support Conversations dataset (ESConv) \citep{Liu2021TowardsES} has marked a significant milestone, providing a rich resource for researchers to delve into and enhance emotional support dialogue systems. Despite this advancement, there remains a notable gap in the state-of-the-art evaluation methods for such systems. Researchers have tried to build and improve systems that either align closely with the gold standard responses in the dataset (responses from Amazon MTurk workers certified as emotional supporters) or focus on enhancing the model's ability to plan subsequent strategies. However, the predominant metric for comparison in these works remains the alignment with these gold standard responses. We argue that this approach may not be the most effective for several reasons. First, in the realm of emotional support, there is often no single 'correct' strategy for continuing a conversation. Second, even when a model bases its response on a specific strategy, there are numerous potential high-quality responses that could be equally effective.

In our research, we adopt a different perspective, reevaluating the core problem in the context of recent advancements. With the advent of Large Language Models (LLMs), generating natural and fluent text has become less of a challenge. Our focus, therefore, shifts to a more nuanced aspect: the degree to which we can effectively guide these LLMs to adhere to specific emotional support strategies during extended conversations, and importantly, \textbf{how we can evaluate and quantify their proficiency in following these strategies}. This approach acknowledges the proficiency of LLMs in text generation while emphasizing the critical need for strategic control and direction in prolonged interactive scenarios. The challenge extends beyond merely directing the conversation, delving into the realm of assessing and quantifying the model’s adherence to the predefined emotional support strategies. Below are the main contributions of our work:

\paragraph{Introducing Strategy Relevant Attention (SRA): Measuring Strategy Adherence in Conversational AI Through Attention Lens} We introduce a novel proxy metric termed \emph{Strategy Relevant Attention (SRA)}, designed to quantitatively assess the extent to which a model aligns its attention with the strategic directives provided in prompts. This metric facilitates the comparative analysis of different models in terms of their efficacy in guiding model adherence to predefined strategies. Furthermore, SRA aids in the development of prompts that enhance the model's ability to maintain strategic consistency throughout prolonged conversations. Through rigorous evaluation, encompassing both automated and human assessments, we establish a significant correlation between a model's adherence to strategy and its SRA score, highlighting the utility of SRA in evaluating a model's strategy following capability.

\paragraph{Release of an Extended ESConv Dataset}
As a practical contribution to the field, we release an extensive synthetic dataset. This dataset, an extension of the existing ESConv dataset, features multiple strategy continuations. It serves as a valuable resource for further research and development in building steerable emotional support agents suitable for long conversations. 

% \paragraph{Propose a Strategy Classifier}
% Using our proposed dataset we have fine-tuned a Roberta large model to classify utterances in a conversation based on the emotional support strategy they follow

\paragraph{A Steerable Emotional Support Model}
In addition, we fine-tune Llama2-7B-chat and Llama3-8B-instruct models under multiple strategic conditions to improve its steerability throughout extended conversations. Our results demonstrate that the fine-tuned models exhibit significantly better steerability in long conversations compared to the base models. This improvement is also consistent with our finding that the fine-tuned models achieve a higher SRA metric.

\section{Related Work}

\subsection{Emotional Support Conversation Systems}

The landscape of Emotional Support (ES) systems has undergone significant evolution, shaped largely by the nature and complexity of datasets available for research. Early ES datasets predominantly consisted of single-turn conversations (\citep{Medeiros2018UsingCF}, \citep{Sharma2020ACA}), leading to a research focus primarily on developing Emotional Support Conversation (ESC) systems for these simplified, single-interaction scenarios (\citep{Sharma2021TowardsFE}, \citep{Hosseini2021ItTT}).This approach, while foundational, did not fully encapsulate the dynamic and multi-faceted nature of real-world emotional support interactions. The release of the first multi-turn ESC dataset, ESConv \citep{Liu2021TowardsES}, marked a pivotal shift in this domain. This dataset opened up new avenues for exploring data-driven approaches in multi-turn ESC systems.

\citep{Peng2022ControlGU} introduced an innovative hierarchical graph network, aiming to effectively utilize both the global emotion cause and the local user intention in emotional support conversations. Moving away from relying on a single strategy for response generation, \citep{Tu2022MISCAM} incorporated commonsense knowledge and a mix of response strategies into the framework of emotional support conversation. \citep{Cheng2022ImprovingME} put forward the concept of look-ahead strategy planning, a method designed to select strategies that could yield the best long-term effects in emotional support dialogue. In a further advancement, \citep{Peng2022FADOFD} explored the selection of appropriate strategies based on the feedback from the conversation seeker. More recently \citep{Zhao2023TransESCSE} addressed the challenge of performing a smooth transition in an utterance level based on semantics, emotions and strategies embedded in each utterance. More closely related to our research, \citep{Zheng2023BuildingES} introduced a synthetic dataset with richer annotations and experimented with fine tuning llama models for this task using parameter efficient methods and showed that it outperforms previous work.

\subsection{Large Language Models' Behavior in Long-Context Scenarios}

The interaction of large language models (LLMs) with long-context scenarios has been a subject of considerable research interest and is particularly relevant to this work. \citep{Krishna2022RankGenIT} observed that in moderately-sized Transformer language models, the quality of neural generation tends to deteriorate when dealing with long contexts. In a study focused on long-context models, \citep{Sun2021DoLL} reported that while extended contexts do enhance the prediction accuracy for a limited set of tokens, the overall improvement remains marginal. Further exploring this domain, \citep{Qin2022TheNT} conducted an analysis on the performance of efficient Transformers across a range of long-context downstream NLP tasks. Their findings reveal a recency bias in long-context Transformers, indicating that these models do not effectively leverage long-range context. In a recent study \citep{Liu2023LostIT} revealed "lost in the middle" effect in SOTA LLM models which indicates that these models can overlook the tokens in the middle of the input. As a subsequent study, researchers showed that instruction fine-tuned versions of these models still overlook the middle and tail of the input prompt, but this happens less than pre-trained models \citep{Wu2023FromLM}.

\subsection{Steering Language Models}

The ability to steer language models (LMs) towards desired behaviors has become increasingly important, particularly for applications requiring alignment with human preferences or specific conversational strategies. Recent research has explored various methods for achieving this steerability, ranging from fine-tuning and prompt engineering to more direct manipulation of model activations. \citep{Turner2023ActivationAS} introduces a lightweight method for model control that directly modifies the activations of LMs during inference. \citep{Alves2023SteeringLL} explores the potential of steering LMs in the domain of machine translation. By leveraging both fine-tuning and in-context learning, this work demonstrates how LMs can be guided to adhere to specific linguistic or stylistic preferences. More relevant to our work, \citet{Dong2023SteerLMAC} proposes a novel approach for steering language models using attribute-conditioned supervised fine-tuning (SFT). By conditioning models on desired attributes during the training process. Our approach, inspired by this work and our proposed attention-based adherence signal (SRA) tries to tackle this steerability challenge in the domain of emotional support conversations.

\section{Preliminaries}

% \subsection{ (zero shot) prompting LLMS ?}
\subsection{ESConv Dataset}
Our research leverages the Emotional Support Conversation dataset, ESConv \citep{Liu2021TowardsES}, which is notably characterized by its inclusion of long conversations, averaging 30 turns per dialogue. This aspect is of paramount importance to our work, as our analysis specifically targets the dynamics of extended dialogues in emotional support contexts. In these interactions, individuals seeking support (seekers) engage with others (supporters) who assist them in navigating through challenging emotional states. The supporters' responsibilities encompass recognizing the seekers' problems, providing consolation, and suggesting actionable solutions to address their concerns according to a predefined strategy. Appendix \ref{sec:appendix-ds-esconv} summarizes the statistics of this dataset and it's key features.

\subsection{Transformers and Auto Regressive Language Models}
Given a sequence of input embeddings $\{e_m\}_{m=1}^{L}$ in $R^d$, where $L$ is the length of the input sequence, a transformer language model with $M$ layers and $H$ attention heads processes each embedding $e_m$. At each layer, the model transforms the embeddings into their corresponding query, key, and value vectors in $R^{d/H}$ as shown in equation \ref{eq:kqv}:

\begin{align}
\label{eq:kqv}
    q_m &= W^q e_m, \nonumber\\
    k_m &= W^k e_m, \nonumber\\
    v_m &= W^v e_m,
\end{align}
    
where $W^q, W^k, W^v \in R^{d/H \times d}$ are learnable matrices. We will then use these vectors to calculate attention weights over previous tokens (equation \ref{eq:lmn}) where $h$ is the corresponding attention head.

\begin{equation}
\label{eq:lmn}
    l^h_{mn} = 
    \begin{cases} 
        \langle q_m^h, k_n^h \rangle, & \text{if } m \geq n, \\
        -\infty, & \text{otherwise},
    \end{cases}
\end{equation}

We will then apply a scaled softmax normalization to calculate the final attention weights $a^h_{m,n}$ as in equation \ref{eq:softmax}
\begin{equation}
\label{eq:softmax}
    a^h_{m,n} = \frac{\exp\left(l^h_{m,n} / \sqrt{d/H}\right)}{\sum_{i=1}^{L} \exp\left(l^h_{m,i} / \sqrt{d/H}\right)}
\end{equation}

The attention weights will be used to calculate the final output embedding $o^h_{m,n}$ at position $m$ for head $h$ (equation \ref{eq:outemb})

\begin{equation}
\label{eq:outemb}
    o^h_{m,n} = \sum_{n=1}^{L} a_{m,n}^{(h)} v_n^{(h)}
\end{equation}

\section{Methodology}

\begin{figure*}[ht]
    \centering
    \includegraphics[width=\textwidth]{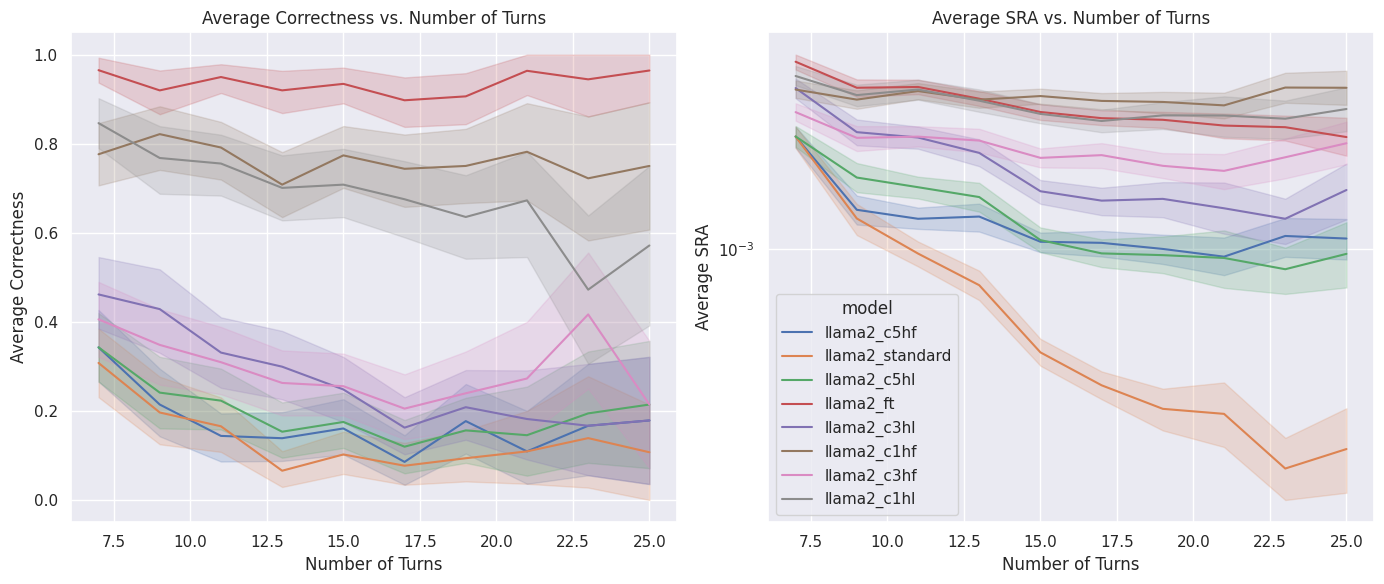} % Adjust the path to your image file and the scaling as needed
    \caption{Left: average accuracy of the strategy following for each model with respect to the turn of the conversation, Right: average SRA of the responses with respect to the turn of the conversation}
    \label{fig:depth} % Label for referencing the figure in text
\end{figure*}

When we attempted to force the model to follow specific strategies using a standard prompt, we noticed a trend: as the conversation extended, the model's responses became increasingly indifferent to the system prompt, particularly to the prompted strategy. Specifically, the model began to generate very general responses, regardless of what the specified strategy was. This tendency to drift towards generic responses irrespective of the strategy input suggests a diminishing sensitivity to the strategic nuances as the dialogue progresses.

Inspired by prior research investigating the impact of token positioning within prompts \citep{Liu2023LostIT}, \citep{Wu2023FromLM}, we formulated a hypothesis concerning the behavior of large language models in extended dialogues. We hypothesize that as the context length increases, the model's attention to tokens related to the prompted strategy decreases. This diminishing focus could result in a drift towards less specific and more generalized responses as the conversation progresses.

To test this hypothesis, \textbf{we introduce the metric "Strategy Relevant Attention (SRA)". This metric is designed to measure the degree to which the tokens generated by the model are focused on the strategy-relevant tokens present in the input}. The core objective is to build a prompting template that consistently maintain attention on the strategic aspects of the dialogue over time. By quantifying the model's adherence to the prompted strategy, this metric serves as a critical tool in assessing the effectiveness of different models in following strategic directions throughout the conversation.

\subsection{Strategy Relevant Attention}
\label{sec:SRA}

Informed by the concept of attention mechanisms, we hypothesise that the level of attention paid to strategy-centric tokens could be a pivotal factor in determining the model's proficiency in adhering to the set strategy, although this remains to be empirically validated. To quantify this assumption, we aggregate the attention weights of the strategy relevant tokens over all heads and all layers for the generated response tokens. 

Let's assume that the strategy relevant tokens span from token $S_b$ to $S_e$ and the response tokens generated by the model span from token $L+1$ to token $L+R$ where $R$ is the length of the response tokens. We can define the attention weight matrix as $A \in R^{M \times H \times R \times L}$ ($M$ being number of attention layers and $H$ being the number of attention heads) in which each element represents the attention of a response token over a prompt token in a specific head and layer of the LLM following the equation \ref{eq:softmax}. Equation \ref{eq:attw} formulates Strategy Relevant Attention ($SRA$) as the aggregate attention of response tokens on the strategy relevant tokens.

\begin{align}
\label{eq:attw}
    SRA^{agg}_{r,l} &= \frac{1}{MH} \sum_{m=1}^{M} \sum_{h=1}^{H} A_{m,h,r,l}, \nonumber \\
    SRA &= \frac{1}{|S_e - S_b| \times R} \sum_{r=1}^{R} \sum_{l=S_b}^{S_e} \bar{SRA}^{agg}_{r,l} \in \mathbb{R}
\end{align}

\begin{table}[ht]
\centering
\resizebox{\columnwidth}{!}{
\begin{tabular}{|l|c|c|}
\hline
\textbf{Model} & \textbf{Accuracy} & \textbf{Log-SRA} \\ \hline
llama3-8b-instruct\_standard & 0.594 & -6.916 \\ 
llama3-8b-instruct\_c5hf & 0.658 & -6.607 \\ 
llama3-8b-instruct\_c3hf & 0.711 & -6.337 \\ 
llama3-8b-instruct\_c5hl & 0.736 & -6.711 \\ 
llama3-8b-instruct\_c3hl & 0.798 & -6.549 \\ 
llama3-8b-instruct\_c1hf & 0.811 & \textbf{-5.758} \\ 
llama3-8b-instruct\_c1hl & 0.821 & -6.082 \\  \hline
llama3-8b-instruct\_ours & \textbf{0.940} & -5.760 \\  \hline
llama2-7b-chat\_standard & 0.144 & -7.294 \\
llama2-7b-chat\_c5hf & 0.178 & -6.802 \\ 
llama2-7b-chat\_c5hl & 0.202 & -6.781 \\ 
llama2-7b-chat\_c3hf & 0.294 & -6.394 \\ 
llama2-7b-chat\_c3hl & 0.295 & -6.504 \\ 
llama2-7b-chat\_c1hl & 0.715 & -6.164 \\ 
llama2-7b-chat\_c1hf & 0.765 & \textbf{-6.112} \\  \hline
llama2-7b-chat\_ours & \textbf{0.933} & -6.126 \\ \hline
\end{tabular}
}
\caption{Log-SRA and strategy following accuracy of our proposed models versus the baselines. For our models we use the standard prompting template.}
\label{tab:log_sra_acc}
\end{table}

\subsection{Prompting Baselines}
\label{sec:method-baselines}

\begin{figure}[!ht]
    \centering
    \includegraphics[width=0.47\textwidth]{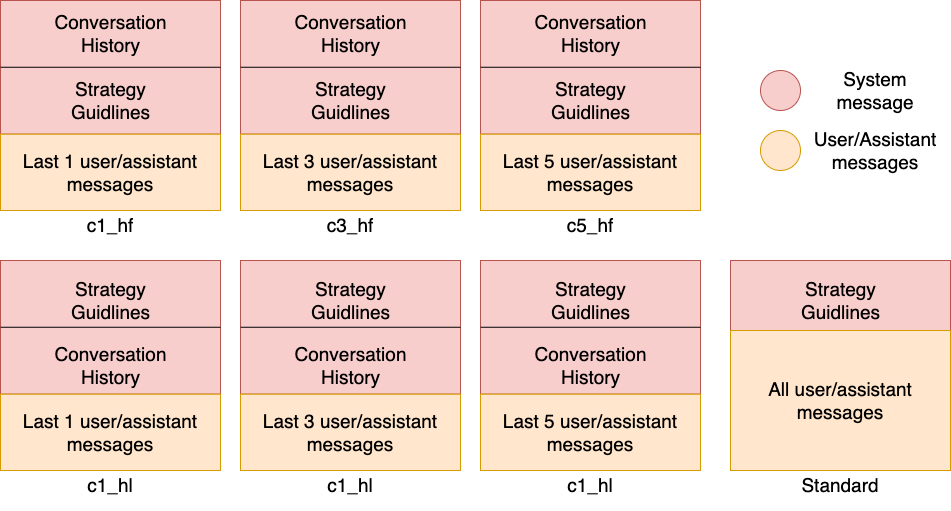} % Adjust the path to your image file and the scaling as needed
    \caption{Six experimental prompt templates to measure SRA with respect to the position of strategy guidelines inside the prompt.}
    \label{fig:prompts} % Label for referencing the figure in text
\end{figure}

For the baseline, we adhered to the standard prompt template as proposed by the Llama model developers \citep{Touvron2023Llama2O}. This involves incorporating the strategy into the system message of the input prompt, followed by the conversation history up to the last message from the emotional support seeker. In contrast, we also design 6 other prompt templates as described in figure \ref{fig:prompts}. These variations include maintaining only 1, 3, or 5 of the most recent messages in the user/assistant message section of the prompt and relocating the remainder of the conversation history to either the beginning or the end of the system message resulting in \emph{c1\_hf, c1\_hl, c3\_hf, c3\_hl, c5\_hf, c5\_hl} templates. This alteration aims to test the impact of prompt structure on the model's attention to strategy guidelines in extended dialogues. To create a follow-up response in the conversation using a particular strategy, we incorporate the \textit{situation} from the original ESConv dataset (it is a short summary of the emotional challenge the help seeker is dealing with), \textit{strategy} which is the strategy that the help seeker is supposed to follow in the next utterance, \textit{strategy description} which is the definition of the strategy\footnote{full list of strategies and their definitions can be found in appendix \ref{sec:appendix-ds-exesconv}}, and all utterances into the prompt template. We then feed the resulting sequence into the model and generate the next utterance.

\subsection{Extended ESConv Dataset}
\label{sec:exesc}

The ESConv dataset initially categorizes the supporter's conversational strategies, identifying eight types, such as questioning, reflecting feelings, and providing suggestions. However, our study seeks to explore the intricacies of emotional support with a more granular approach. Taking inspiration from the study by \citep{Zheng2023BuildingES} which developed a more detailed method for categorizing support strategies, we have decided to use this advanced classification in our study. We've detailed each strategy along with a description of the strategy and more details about this dataset in appendix \ref{sec:appendix-ds-exesconv}. Using these new categories, \textbf{we expanded the ESConv dataset into several variations for just one turn conditioned on multiple strategies}. We picked a random conversation from the dataset and split it at a random point between the 5th and 23th turn. We chose these points to make sure we continued the conversation in the most appropriate spots. For instance, it wouldn't make sense to start \emph{Collaborative Planning} when someone is just saying goodbye, or to use \emph{Reflective Statement} when just greeting. We always split the conversation after the person seeking help has spoken, allowing the model to take over as the supporter. Then, with a specific model and a prompting template, \textbf{we carried the conversation forward by one turn} using some of the 15 support strategies \citep{Zheng2023BuildingES} mentioned. This creates various strategy conditioned single-turn continuations of the conversations. However, we couldn't try out every single combination because of computing constraints. We hold out 100 conversations for testing different hypothesises. Table \ref{tab:stat-extes} shows the statistics of our proposed extended dataset.

\begin{table}
    \centering
    \resizebox{\columnwidth}{!}{
    
    \begin{tabular}{|c|c|}
        \hline
        Number of dialogs & 1,297\\ \hline
        \makecell{Number of strategy conditioned\\continuations} & 41,822\\ \hline
        \makecell{min conversation history length} & 5\\ \hline
        \makecell{max conversation history length} & 23 \\ \hline
        \makecell{avg conversation history length} & 11.76\\ \hline
         train/test/validation split & 1147 / 100 / 50 conversations\\ \hline
         train/test/validation examples & 36,923 / 3,292 / 1,607\\ \hline
    \end{tabular}
    }
    \caption{Statistics of our proposed extended ESconv dataset}
    \label{tab:stat-extes}
\end{table}

\subsection{Training a Steerable Model}
\label{sec:lora}
Informed by our proposed SRA metric, we selected the prompting baseline with the highest SRA (c1\_hf) and used llama2-13b-chat model to generate the Extended ESConv dataset, as discussed in \ref{sec:exesc}. We fine-tuned \emph{Llama2-7b-chat} and \emph{Llama3-8b-instruct} models using the LoRA method \cite{Hu2021LoRALA}, focusing exclusively on the last utterance's strategy-conditioned continuations within the standard prompting setup. This approach enables the model to prioritize the system prompt while generating new utterances. Our objective is to enhance the model's attention to the system prompt, thereby increasing its steerability in this setup. We fine-tuned the model on a single A100 GPU, utilizing the default training configurations of base models with a cumulative batch size of 64 and a cosine learning rate schedule with hard restarts for 5 epochs. We mask the conversation history for loss calculation and only measure the negative log-likelihood on the last utterance. Also, followed by the convention set by the LoRA paper we used alpha of 256 twice the size of rank 128 for the adaptors.

\subsection{Strategy Classifier}
\label{sec:eval_cls}

We utilize the same dataset provided in \ref{sec:lora} to train a RoBERTa-large sequence classifier, designed to categorize the strategy employed in a single response. The model is trained on pairs of response and prompted strategy from our best performing prompting baseline. Note that this is a weakly labeled dataset as the responses might not well follow the prompted strategy. Based on human annotation on a held-out dataset of 1000 utterance samples from our test set, the trained classifier achieves an accuracy of 93.6\%. More details on the performance of this model and it's error analysis can be found in appendix \ref{appendix:classifer}. This classifier is trained to automatically assess the strategy adherence of different models. The model is trained for 5 epochs using a batch size of 128 on a single A100 GPU. We also perform an extensive qualitative and quantitative analysis on the predictability of different model responses in appendix \ref{sec:appendix_pred}.

\section{Experimental Setup and Results}

As discussed in \ref{sec:exesc}, we hold out 100 conversations for experiments and tests. We perform the experiments discussed in this section on these 100 conversations which haven't been seen by our fine-tuned models during training. For generating responses we set the decoding strategy to sampling with \emph{top\_p} of 0.9 and \emph{temperature} of 0.7 for all of the models.

\begin{figure}[ht]
    \centering
    \includegraphics[width=\columnwidth]{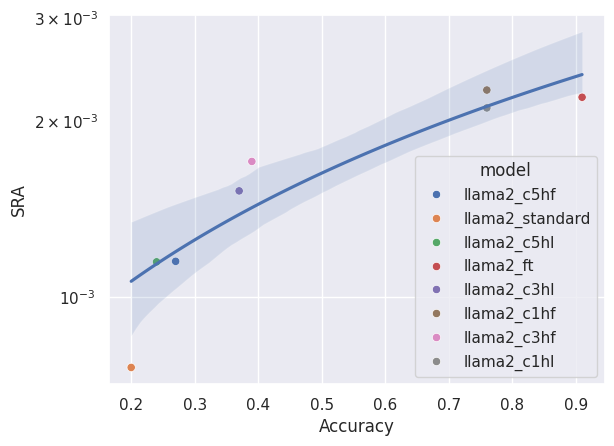} % Adjust the path to your image file and the scaling as needed
    \caption{Correlation of the SRA metric with the accuracy of the strategy following for llama2-7b-chat fine-tuned model vs. baselines. Y-axis is in logarithmic scale.}
    \label{fig:corr} % Label for referencing the figure in text
\end{figure}

\subsection{How does SRA correlate with strategy adherence?}
\label{sec:eval_sra}

In this section, we aim to test our hypothesis regarding the correlation between the SRA metric and the steerability of the model in strategy adherence. Specifically,\textbf{ we hypothesize that the SRA metric quantifies the extent to which a model, identical in parameters to another, allocates attention to strategy-specific tokens, thereby indicating its proficiency in adhering to the intended strategy.} We follow the approach outlined in section \ref{sec:exesc} , use llama2-7b-chat and extend the test set conversations using all of the prompting baselines. Note that for our fine-tuned model we also use the standard prompt template.
To test this hypothesis, we utilize the strategy classifier trained in \ref{sec:eval_cls}
to evaluate the strategies employed in the generated responses by each model. We compare our fine-tuned model against the baseline prompts given the same model size (llama2-7b-chat). Subsequently, we measure how closely the accuracy of strategy adherence correlates with the SRA metric. Figure \ref{fig:corr} illustrates the results. We observe that the SRA metric has a Pearson correlation of 0.94 with the correctness of the strategies followed by the models, indicating a significant correlation. We also perform the same experiment using llama3-8b-instruct and observed the same strong correlation between strategy adherence and SRA of 0.8.

\subsection{How does steerability of the model change in depth of the conversations?}
\label{sec:eval_long}

We use the strategy classifier in \ref{sec:eval_cls} to compare the accuracy of the strategy adherence of our proposed models versus the baselines. We compare our fine-tuned llama3-8b-instruct and llama2-7b-chat with baselines explained in section \ref{sec:method-baselines} over both SRA metric and strategy adherence accuracy. Note again that for fine-tuned models we use the standard prompting. We summarize the results in table \ref{tab:log_sra_acc}. Our fine-tuned models consistently outperforms the baselines with an \textbf{improvement of 78.9\% over the llama-2 base model and 37.6\%} over llama-3 base model in strategy adherence accuracy and gains significantly more SRA after fine-tuning. It is also worth mentioning that llama3-8b-instruct model is significantly improved over llama2-7b-chat in steerability as it outperforms it by 45\%. 

Figure \ref{fig:depth} illustrates the amount of SRA and average accuracy of the strategy following with respect to the turn of the conversation for llama2-7b-chat and it's corresponding baselines. We also provide the results for the same set of experiments over llama3-8b-instruct in appendix \ref{appendix:llama3}. We observe that the baseline prompts informed by our SRA metric are able to maintain a high SRA and strategy adherence as we go deeper into the conversation and our proposed model outperforms the baselines and maintains a robust SRA and adherence throughout the conversation. Also the \emph{c1\_hf} prompt achieves the most stable strategy following behavior compared to the other baselines. More importantly, we observe that our fine-tuned model, despite maintaining the same prompting template as the standard baseline, significantly achieves higher SRA and strategy following accuracy during the conversation.

\subsection{Does better strategy following deteriorate conversationality of the model?}

To this end, we found out that we can improve strategy adherence or steerability of the model by designing more efficient prompts or fine-tuning the model with synthetic data to enforce the model to attend more to the strategy directions. However, \textbf{it is important to evaluate other dimensions of performance and make sure that the resulting models are not less coherent, natural and consistent.} Note that since we are forcing the models to blindly follow different strategies at different stages of the conversation, we can't simply compare the helpfulness of two model responses as the helpfulness is not the relevant metric here since the followed strategy might not be optimal. Therefore, aside from the strategy following capability we try to measure coherence, naturalness and the quality of the responses.

\subsubsection{Model based evaluation results}
\label{sec:modelbasedeval}

We compare our fine-tuned model's responses with it's initial base model before fine-tuning. We generate 500 random strategy-conditioned utterances following the same approach as section \ref{sec:exesc} from the held out 100 conversations using llama2-7b-chat, llama3-8b-instruct and their fine-tuned variations using standard prompting. 
We use gpt-4o as the judge LLM. The prompt template along with some examples are shown in appendix \ref{appendix:model-based-eval}. We also mitigate possible positional bias discussed in \cite{Zheng2023JudgingLW} and alternate between model responses and call it a tie in case of mismatched judges. Table \ref{tab:h2h} shows the wining rate of our proposed llama-2 and llama-3 fine-tuned models versus the initial models. Both of our llama-2 and llama-3 models significantly win over baselines. This highlights that our approach not only enhances the strategy adherence of the base models, but also maintains consistency, coherence and quality.

\begin{table}
    \centering
    \resizebox{\columnwidth}{!}{
    \begin{tabular}{c|ccc}
          & Win(\%) & Tie(\%) & Lose(\%)\\
          \hline
         \makecell{Llama2-7b-chat\\(ours vs standard)} & \textbf{74.12} & 18.63 & 7.25\\
         \hline
         \makecell{Llama3-8b-instruct\\(ours vs standard)} & \textbf{50.78} & 35.29 & 13.92\\
    \end{tabular}

    }
    \caption{Head to head comparison of our fine-tuned models with their base models.}
    \label{tab:h2h}
\end{table}

\subsubsection{Human evaluation results}

\begin{figure}[!ht]
    \centering
    \includegraphics[width=0.47\textwidth]{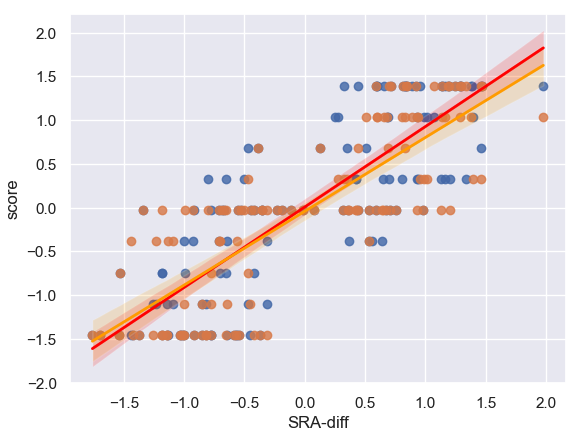} % Adjust the path to your image file and the scaling as needed
    \caption{y-axis shows the normalized score of the annotators for each annotation task and x-axis shows the normalized log of difference between responses in the annotation task. }
    \label{fig:human-eval} % Label for referencing the figure in text
\end{figure}

In addition to model based analyses, we also ask human annotators to score how good different models adhere to strategies while staying coherent, consistent and natural. We generate responses to a given conversation history using two distinct models picked among \emph{c1\_hf}, \emph{c3\_hf}, \emph{standard} and our fine-tuned version with standard prompting and use llama2-7b-chat as the assistant. Same as section \ref{sec:modelbasedeval}, we generate 150 head to head strategy-conditioned continuation comparisons. We then compute the SRA for both responses, which serves as a preliminary quantitative measure of strategic alignment. Subsequently, two human annotators are tasked with evaluating the responses, assigning scores based on the perceived effectiveness of each response in following the outlined strategy and being coherent, consistent and natural. Finally, we measure the Pearson correlation between the human score and the difference between SRA metrics of the two responses. Details of the annotation task are explained in appendix \ref{appendix-annotation}. As depicted in figure \ref{fig:human-eval}, we observe a high Pearson correlation of 0.80 and 0.82 between the each of the annotators' scores and the difference in SRA for the two responses with Krippendorff’s Alpha agreement of 0.91 between the annotators. This result, highlights the effectiveness of our proposed SRA metric in measuring the adherence of the models to the prompted strategy. We also observe a 79.41\% win-rate of our fine-tuned model versus other baselines.

\section{Conclusion}

In this study, we addressed the challenges that large language models (LLMs) face when engaging in long-form conversational setups, specifically focusing on maintaining consistent strategy adherence in emotional support conversations. We introduced the Strategy-Relevant Attention (SRA) metric to monitor how effectively models attend to strategic directives throughout interactions. Our findings show that the Llama-2 and Llama-3 models struggle to consistently follow the intended strategies, particularly as conversations progress.

Leveraging the insights from the SRA metric, we explored various prompting techniques to enhance strategy adherence and constructed a synthetic, strategy-conditioned extension of the ESConv dataset. Our fine-tuned models, trained on this extended dataset, demonstrated significant improvements in steerability or strategy adherence compared to baseline models.

Furthermore, through model-based and human evaluations, we validated that our fine-tuned models not only excel in maintaining strategic focus but also preserve key conversational qualities such as naturalness, relevance, consistency, and coherence. These results highlight the potential of our approach in steering LLMs more effectively in emotionally supportive dialogues and set the stage for future work in refining long-horizon conversational planning.

\section{Limitations}

While our research on the Strategy-Relevant Attention (SRA) metric demonstrates significant advancements in conversational AI, it is not without limitations. Firstly, the generalizability of SRA across diverse LLM architectures and configurations remains to be fully explored. Additionally, the effectiveness of SRA in scenarios beyond emotional support conversations, especially in more complex or nuanced interactions, requires further investigation. Also, in this work we only focus on the ability of these models for following strategy. Although this is an important skill in a conversational agent, but we do not study the helpfulness or effectiveness of different strategies at different stages of the conversation and leave it to the future work.

\section{Ethical Considerations}

Given that the LLMs are designed to engage in emotionally sensitive conversations, it is crucial to ensure user privacy and confidentiality. All data used for training and evaluation should be anonymized, with personal identifiers removed. This is mitigated by the authors of the ESConv dataset and since we are extending the same dataset, we do not encounter such challenges. Also, since the model engages with users seeking emotional support, it must be carefully designed to avoid causing harm. It is essential to ensure that the model's responses are empathetic, supportive, and non-triggering. Continuous monitoring, human-in-the-loop oversight, and the inclusion of mental health professionals in the evaluation process are necessary to maintain the emotional safety of users. This work only focuses on strategy adherence and does not consider the planning aspect of the emotional support task. It is also important that the users interacting with the model be made aware that they are engaging with an AI system rather than a human. Clear communication about the system's capabilities, limitations, and the nature of the support it can provide is essential. Informed consent should be obtained, ensuring users understand the model's role, data usage policies, and the scope of emotional support it offers.

% \subsection{References}

% \nocite{Ando2005,andrew2007scalable,rasooli-tetrault-2015}

% The \LaTeX{} and Bib\TeX{} style files provided roughly follow the American Psychological Association format.
% If your own bib file is named \texttt{custom.bib}, then placing the following before any appendices in your \LaTeX{} file will generate the references section for you:
% \begin{quote}
% \begin{verbatim}
% \bibliography{custom}
% \end{verbatim}
% \end{quote}

\bibliography{custom}

\begin{thebibliography}{24}
\providecommand{\natexlab}[1]{#1}

\bibitem[{Alves et~al.(2023)Alves, Guerreiro, Alves, Pombal, Rei, de~Souza, Colombo, and Martins}]{Alves2023SteeringLL}
Duarte~M. Alves, Nuno~M. Guerreiro, Joao Alves, Jos{\'e}~P. Pombal, Ricardo Rei, Jos'e G.~C. de~Souza, Pierre Colombo, and Andr{\'e} Martins. 2023.
\newblock \href {https://api.semanticscholar.org/CorpusID:264405904} {Steering large language models for machine translation with finetuning and in-context learning}.
\newblock \emph{ArXiv}, abs/2310.13448.

\bibitem[{Cheng et~al.(2022)Cheng, Liu, Li, Wang, Zhao, Liu, Liang, and Zheng}]{Cheng2022ImprovingME}
Yi~Cheng, Wenge Liu, Wenjie Li, Jiashuo Wang, Ruihui Zhao, Bang Liu, Xiaodan Liang, and Yefeng Zheng. 2022.
\newblock \href {https://api.semanticscholar.org/CorpusID:252780132} {Improving multi-turn emotional support dialogue generation with lookahead strategy planning}.
\newblock In \emph{Conference on Empirical Methods in Natural Language Processing}.

\bibitem[{Dong et~al.(2023)Dong, Wang, Sreedhar, Wu, and Kuchaiev}]{Dong2023SteerLMAC}
Yi~Dong, Zhilin Wang, Makesh~Narsimhan Sreedhar, Xianchao Wu, and Oleksii Kuchaiev. 2023.
\newblock \href {https://api.semanticscholar.org/CorpusID:263830508} {Steerlm: Attribute conditioned sft as an (user-steerable) alternative to rlhf}.
\newblock \emph{ArXiv}, abs/2310.05344.

\bibitem[{Dubey et~al.(2024)Dubey, Jauhri, Pandey, Kadian, Al-Dahle, Letman, Mathur, Schelten, Yang, Fan, Goyal, Hartshorn, Yang, Mitra, Sravankumar, Korenev, Hinsvark, Rao, Zhang, Rodriguez, Gregerson, Spataru, Roziere, Biron, Tang, Chern, Caucheteux, Nayak, Bi, Marra, McConnell, Keller, Touret, Wu, Wong, Ferrer, Nikolaidis, Allonsius, Song, Pintz, Livshits, Esiobu, Choudhary, Mahajan, Garcia-Olano, Perino, Hupkes, Lakomkin, AlBadawy, Lobanova, Dinan, Smith, Radenovic, Zhang, Synnaeve, Lee, Anderson, Nail, Mialon, Pang, Cucurell, Nguyen, Korevaar, Xu, Touvron, Zarov, Ibarra, Kloumann, Misra, Evtimov, Copet, Lee, Geffert, Vranes, Park, Mahadeokar, Shah, van~der Linde, Billock, Hong, Lee, Fu, Chi, Huang, Liu, Wang, Yu, Bitton, Spisak, Park, Rocca, Johnstun, Saxe, Jia, Alwala, Upasani, Plawiak, Li, Heafield, Stone, El-Arini, Iyer, Malik, Chiu, Bhalla, Rantala-Yeary, van~der Maaten, Chen, Tan, Jenkins, Martin, Madaan, Malo, Blecher, Landzaat, de~Oliveira, Muzzi, Pasupuleti, Singh, Paluri, Kardas, Oldham, Rita,
  Pavlova, Kambadur, Lewis, Si, Singh, Hassan, Goyal, Torabi, Bashlykov, Bogoychev, Chatterji, Duchenne, Çelebi, Alrassy, Zhang, Li, Vasic, Weng, Bhargava, Dubal, Krishnan, Koura, Xu, He, Dong, Srinivasan, Ganapathy, Calderer, Cabral, Stojnic, Raileanu, Girdhar, Patel, Sauvestre, Polidoro, Sumbaly, Taylor, Silva, Hou, Wang, Hosseini, Chennabasappa, Singh, Bell, Kim, Edunov, Nie, Narang, Raparthy, Shen, Wan, Bhosale, Zhang, Vandenhende, Batra, Whitman, Sootla, Collot, Gururangan, Borodinsky, Herman, Fowler, Sheasha, Georgiou, Scialom, Speckbacher, Mihaylov, Xiao, Karn, Goswami, Gupta, Ramanathan, Kerkez, Gonguet, Do, Vogeti, Petrovic, Chu, Xiong, Fu, Meers, Martinet, Wang, Tan, Xie, Jia, Wang, Goldschlag, Gaur, Babaei, Wen, Song, Zhang, Li, Mao, Coudert, Yan, Chen, Papakipos, Singh, Grattafiori, Jain, Kelsey, Shajnfeld, Gangidi, Victoria, Goldstand, Menon, Sharma, Boesenberg, Vaughan, Baevski, Feinstein, Kallet, Sangani, Yunus, Lupu, Alvarado, Caples, Gu, Ho, Poulton, Ryan, Ramchandani, Franco, Saraf,
  Chowdhury, Gabriel, Bharambe, Eisenman, Yazdan, James, Maurer, Leonhardi, Huang, Loyd, Paola, Paranjape, Liu, Wu, Ni, Hancock, Wasti, Spence, Stojkovic, Gamido, Montalvo, Parker, Burton, Mejia, Wang, Kim, Zhou, Hu, Chu, Cai, Tindal, Feichtenhofer, Civin, Beaty, Kreymer, Li, Wyatt, Adkins, Xu, Testuggine, David, Parikh, Liskovich, Foss, Wang, Le, Holland, Dowling, Jamil, Montgomery, Presani, Hahn, Wood, Brinkman, Arcaute, Dunbar, Smothers, Sun, Kreuk, Tian, Ozgenel, Caggioni, Guzmán, Kanayet, Seide, Florez, Schwarz, Badeer, Swee, Halpern, Thattai, Herman, Sizov, Guangyi, Zhang, Lakshminarayanan, Shojanazeri, Zou, Wang, Zha, Habeeb, Rudolph, Suk, Aspegren, Goldman, Damlaj, Molybog, Tufanov, Veliche, Gat, Weissman, Geboski, Kohli, Asher, Gaya, Marcus, Tang, Chan, Zhen, Reizenstein, Teboul, Zhong, Jin, Yang, Cummings, Carvill, Shepard, McPhie, Torres, Ginsburg, Wang, Wu, U, Saxena, Prasad, Khandelwal, Zand, Matosich, Veeraraghavan, Michelena, Li, Huang, Chawla, Lakhotia, Huang, Chen, Garg, A, Silva, Bell,
  Zhang, Guo, Yu, Moshkovich, Wehrstedt, Khabsa, Avalani, Bhatt, Tsimpoukelli, Mankus, Hasson, Lennie, Reso, Groshev, Naumov, Lathi, Keneally, Seltzer, Valko, Restrepo, Patel, Vyatskov, Samvelyan, Clark, Macey, Wang, Hermoso, Metanat, Rastegari, Bansal, Santhanam, Parks, White, Bawa, Singhal, Egebo, Usunier, Laptev, Dong, Zhang, Cheng, Chernoguz, Hart, Salpekar, Kalinli, Kent, Parekh, Saab, Balaji, Rittner, Bontrager, Roux, Dollar, Zvyagina, Ratanchandani, Yuvraj, Liang, Alao, Rodriguez, Ayub, Murthy, Nayani, Mitra, Li, Hogan, Battey, Wang, Maheswari, Howes, Rinott, Bondu, Datta, Chugh, Hunt, Dhillon, Sidorov, Pan, Verma, Yamamoto, Ramaswamy, Lindsay, Lindsay, Feng, Lin, Zha, Shankar, Zhang, Zhang, Wang, Agarwal, Sajuyigbe, Chintala, Max, Chen, Kehoe, Satterfield, Govindaprasad, Gupta, Cho, Virk, Subramanian, Choudhury, Goldman, Remez, Glaser, Best, Kohler, Robinson, Li, Zhang, Matthews, Chou, Shaked, Vontimitta, Ajayi, Montanez, Mohan, Kumar, Mangla, Albiero, Ionescu, Poenaru, Mihailescu, Ivanov, Li, Wang,
  Jiang, Bouaziz, Constable, Tang, Wang, Wu, Wang, Xia, Wu, Gao, Chen, Hu, Jia, Qi, Li, Zhang, Zhang, Adi, Nam, Yu, Wang, Hao, Qian, He, Rait, DeVito, Rosnbrick, Wen, Yang, and Zhao}]{dubey2024llama3herdmodels}
Abhimanyu Dubey, Abhinav Jauhri, Abhinav Pandey, Abhishek Kadian, Ahmad Al-Dahle, Aiesha Letman, Akhil Mathur, Alan Schelten, Amy Yang, Angela Fan, Anirudh Goyal, Anthony Hartshorn, Aobo Yang, Archi Mitra, Archie Sravankumar, Artem Korenev, Arthur Hinsvark, Arun Rao, Aston Zhang, Aurelien Rodriguez, Austen Gregerson, Ava Spataru, Baptiste Roziere, Bethany Biron, Binh Tang, Bobbie Chern, Charlotte Caucheteux, Chaya Nayak, Chloe Bi, Chris Marra, Chris McConnell, Christian Keller, Christophe Touret, Chunyang Wu, Corinne Wong, Cristian~Canton Ferrer, Cyrus Nikolaidis, Damien Allonsius, Daniel Song, Danielle Pintz, Danny Livshits, David Esiobu, Dhruv Choudhary, Dhruv Mahajan, Diego Garcia-Olano, Diego Perino, Dieuwke Hupkes, Egor Lakomkin, Ehab AlBadawy, Elina Lobanova, Emily Dinan, Eric~Michael Smith, Filip Radenovic, Frank Zhang, Gabriel Synnaeve, Gabrielle Lee, Georgia~Lewis Anderson, Graeme Nail, Gregoire Mialon, Guan Pang, Guillem Cucurell, Hailey Nguyen, Hannah Korevaar, Hu~Xu, Hugo Touvron, Iliyan Zarov,
  Imanol~Arrieta Ibarra, Isabel Kloumann, Ishan Misra, Ivan Evtimov, Jade Copet, Jaewon Lee, Jan Geffert, Jana Vranes, Jason Park, Jay Mahadeokar, Jeet Shah, Jelmer van~der Linde, Jennifer Billock, Jenny Hong, Jenya Lee, Jeremy Fu, Jianfeng Chi, Jianyu Huang, Jiawen Liu, Jie Wang, Jiecao Yu, Joanna Bitton, Joe Spisak, Jongsoo Park, Joseph Rocca, Joshua Johnstun, Joshua Saxe, Junteng Jia, Kalyan~Vasuden Alwala, Kartikeya Upasani, Kate Plawiak, Ke~Li, Kenneth Heafield, Kevin Stone, Khalid El-Arini, Krithika Iyer, Kshitiz Malik, Kuenley Chiu, Kunal Bhalla, Lauren Rantala-Yeary, Laurens van~der Maaten, Lawrence Chen, Liang Tan, Liz Jenkins, Louis Martin, Lovish Madaan, Lubo Malo, Lukas Blecher, Lukas Landzaat, Luke de~Oliveira, Madeline Muzzi, Mahesh Pasupuleti, Mannat Singh, Manohar Paluri, Marcin Kardas, Mathew Oldham, Mathieu Rita, Maya Pavlova, Melanie Kambadur, Mike Lewis, Min Si, Mitesh~Kumar Singh, Mona Hassan, Naman Goyal, Narjes Torabi, Nikolay Bashlykov, Nikolay Bogoychev, Niladri Chatterji, Olivier
  Duchenne, Onur Çelebi, Patrick Alrassy, Pengchuan Zhang, Pengwei Li, Petar Vasic, Peter Weng, Prajjwal Bhargava, Pratik Dubal, Praveen Krishnan, Punit~Singh Koura, Puxin Xu, Qing He, Qingxiao Dong, Ragavan Srinivasan, Raj Ganapathy, Ramon Calderer, Ricardo~Silveira Cabral, Robert Stojnic, Roberta Raileanu, Rohit Girdhar, Rohit Patel, Romain Sauvestre, Ronnie Polidoro, Roshan Sumbaly, Ross Taylor, Ruan Silva, Rui Hou, Rui Wang, Saghar Hosseini, Sahana Chennabasappa, Sanjay Singh, Sean Bell, Seohyun~Sonia Kim, Sergey Edunov, Shaoliang Nie, Sharan Narang, Sharath Raparthy, Sheng Shen, Shengye Wan, Shruti Bhosale, Shun Zhang, Simon Vandenhende, Soumya Batra, Spencer Whitman, Sten Sootla, Stephane Collot, Suchin Gururangan, Sydney Borodinsky, Tamar Herman, Tara Fowler, Tarek Sheasha, Thomas Georgiou, Thomas Scialom, Tobias Speckbacher, Todor Mihaylov, Tong Xiao, Ujjwal Karn, Vedanuj Goswami, Vibhor Gupta, Vignesh Ramanathan, Viktor Kerkez, Vincent Gonguet, Virginie Do, Vish Vogeti, Vladan Petrovic, Weiwei Chu,
  Wenhan Xiong, Wenyin Fu, Whitney Meers, Xavier Martinet, Xiaodong Wang, Xiaoqing~Ellen Tan, Xinfeng Xie, Xuchao Jia, Xuewei Wang, Yaelle Goldschlag, Yashesh Gaur, Yasmine Babaei, Yi~Wen, Yiwen Song, Yuchen Zhang, Yue Li, Yuning Mao, Zacharie~Delpierre Coudert, Zheng Yan, Zhengxing Chen, Zoe Papakipos, Aaditya Singh, Aaron Grattafiori, Abha Jain, Adam Kelsey, Adam Shajnfeld, Adithya Gangidi, Adolfo Victoria, Ahuva Goldstand, Ajay Menon, Ajay Sharma, Alex Boesenberg, Alex Vaughan, Alexei Baevski, Allie Feinstein, Amanda Kallet, Amit Sangani, Anam Yunus, Andrei Lupu, Andres Alvarado, Andrew Caples, Andrew Gu, Andrew Ho, Andrew Poulton, Andrew Ryan, Ankit Ramchandani, Annie Franco, Aparajita Saraf, Arkabandhu Chowdhury, Ashley Gabriel, Ashwin Bharambe, Assaf Eisenman, Azadeh Yazdan, Beau James, Ben Maurer, Benjamin Leonhardi, Bernie Huang, Beth Loyd, Beto~De Paola, Bhargavi Paranjape, Bing Liu, Bo~Wu, Boyu Ni, Braden Hancock, Bram Wasti, Brandon Spence, Brani Stojkovic, Brian Gamido, Britt Montalvo, Carl
  Parker, Carly Burton, Catalina Mejia, Changhan Wang, Changkyu Kim, Chao Zhou, Chester Hu, Ching-Hsiang Chu, Chris Cai, Chris Tindal, Christoph Feichtenhofer, Damon Civin, Dana Beaty, Daniel Kreymer, Daniel Li, Danny Wyatt, David Adkins, David Xu, Davide Testuggine, Delia David, Devi Parikh, Diana Liskovich, Didem Foss, Dingkang Wang, Duc Le, Dustin Holland, Edward Dowling, Eissa Jamil, Elaine Montgomery, Eleonora Presani, Emily Hahn, Emily Wood, Erik Brinkman, Esteban Arcaute, Evan Dunbar, Evan Smothers, Fei Sun, Felix Kreuk, Feng Tian, Firat Ozgenel, Francesco Caggioni, Francisco Guzmán, Frank Kanayet, Frank Seide, Gabriela~Medina Florez, Gabriella Schwarz, Gada Badeer, Georgia Swee, Gil Halpern, Govind Thattai, Grant Herman, Grigory Sizov, Guangyi, Zhang, Guna Lakshminarayanan, Hamid Shojanazeri, Han Zou, Hannah Wang, Hanwen Zha, Haroun Habeeb, Harrison Rudolph, Helen Suk, Henry Aspegren, Hunter Goldman, Ibrahim Damlaj, Igor Molybog, Igor Tufanov, Irina-Elena Veliche, Itai Gat, Jake Weissman, James
  Geboski, James Kohli, Japhet Asher, Jean-Baptiste Gaya, Jeff Marcus, Jeff Tang, Jennifer Chan, Jenny Zhen, Jeremy Reizenstein, Jeremy Teboul, Jessica Zhong, Jian Jin, Jingyi Yang, Joe Cummings, Jon Carvill, Jon Shepard, Jonathan McPhie, Jonathan Torres, Josh Ginsburg, Junjie Wang, Kai Wu, Kam~Hou U, Karan Saxena, Karthik Prasad, Kartikay Khandelwal, Katayoun Zand, Kathy Matosich, Kaushik Veeraraghavan, Kelly Michelena, Keqian Li, Kun Huang, Kunal Chawla, Kushal Lakhotia, Kyle Huang, Lailin Chen, Lakshya Garg, Lavender A, Leandro Silva, Lee Bell, Lei Zhang, Liangpeng Guo, Licheng Yu, Liron Moshkovich, Luca Wehrstedt, Madian Khabsa, Manav Avalani, Manish Bhatt, Maria Tsimpoukelli, Martynas Mankus, Matan Hasson, Matthew Lennie, Matthias Reso, Maxim Groshev, Maxim Naumov, Maya Lathi, Meghan Keneally, Michael~L. Seltzer, Michal Valko, Michelle Restrepo, Mihir Patel, Mik Vyatskov, Mikayel Samvelyan, Mike Clark, Mike Macey, Mike Wang, Miquel~Jubert Hermoso, Mo~Metanat, Mohammad Rastegari, Munish Bansal, Nandhini
  Santhanam, Natascha Parks, Natasha White, Navyata Bawa, Nayan Singhal, Nick Egebo, Nicolas Usunier, Nikolay~Pavlovich Laptev, Ning Dong, Ning Zhang, Norman Cheng, Oleg Chernoguz, Olivia Hart, Omkar Salpekar, Ozlem Kalinli, Parkin Kent, Parth Parekh, Paul Saab, Pavan Balaji, Pedro Rittner, Philip Bontrager, Pierre Roux, Piotr Dollar, Polina Zvyagina, Prashant Ratanchandani, Pritish Yuvraj, Qian Liang, Rachad Alao, Rachel Rodriguez, Rafi Ayub, Raghotham Murthy, Raghu Nayani, Rahul Mitra, Raymond Li, Rebekkah Hogan, Robin Battey, Rocky Wang, Rohan Maheswari, Russ Howes, Ruty Rinott, Sai~Jayesh Bondu, Samyak Datta, Sara Chugh, Sara Hunt, Sargun Dhillon, Sasha Sidorov, Satadru Pan, Saurabh Verma, Seiji Yamamoto, Sharadh Ramaswamy, Shaun Lindsay, Shaun Lindsay, Sheng Feng, Shenghao Lin, Shengxin~Cindy Zha, Shiva Shankar, Shuqiang Zhang, Shuqiang Zhang, Sinong Wang, Sneha Agarwal, Soji Sajuyigbe, Soumith Chintala, Stephanie Max, Stephen Chen, Steve Kehoe, Steve Satterfield, Sudarshan Govindaprasad, Sumit Gupta,
  Sungmin Cho, Sunny Virk, Suraj Subramanian, Sy~Choudhury, Sydney Goldman, Tal Remez, Tamar Glaser, Tamara Best, Thilo Kohler, Thomas Robinson, Tianhe Li, Tianjun Zhang, Tim Matthews, Timothy Chou, Tzook Shaked, Varun Vontimitta, Victoria Ajayi, Victoria Montanez, Vijai Mohan, Vinay~Satish Kumar, Vishal Mangla, Vítor Albiero, Vlad Ionescu, Vlad Poenaru, Vlad~Tiberiu Mihailescu, Vladimir Ivanov, Wei Li, Wenchen Wang, Wenwen Jiang, Wes Bouaziz, Will Constable, Xiaocheng Tang, Xiaofang Wang, Xiaojian Wu, Xiaolan Wang, Xide Xia, Xilun Wu, Xinbo Gao, Yanjun Chen, Ye~Hu, Ye~Jia, Ye~Qi, Yenda Li, Yilin Zhang, Ying Zhang, Yossi Adi, Youngjin Nam, Yu, Wang, Yuchen Hao, Yundi Qian, Yuzi He, Zach Rait, Zachary DeVito, Zef Rosnbrick, Zhaoduo Wen, Zhenyu Yang, and Zhiwei Zhao. 2024.
\newblock \href {https://arxiv.org/abs/2407.21783} {The llama 3 herd of models}.
\newblock \emph{Preprint}, arXiv:2407.21783.

\bibitem[{Hosseini and Caragea(2021)}]{Hosseini2021ItTT}
Mahshid Hosseini and Cornelia Caragea. 2021.
\newblock \href {https://api.semanticscholar.org/CorpusID:233238086} {It takes two to empathize: One to seek and one to provide}.
\newblock In \emph{AAAI Conference on Artificial Intelligence}.

\bibitem[{Hu et~al.(2021)Hu, Shen, Wallis, Allen-Zhu, Li, Wang, and Chen}]{Hu2021LoRALA}
J.~Edward Hu, Yelong Shen, Phillip Wallis, Zeyuan Allen-Zhu, Yuanzhi Li, Shean Wang, and Weizhu Chen. 2021.
\newblock \href {https://api.semanticscholar.org/CorpusID:235458009} {Lora: Low-rank adaptation of large language models}.
\newblock \emph{ArXiv}, abs/2106.09685.

\bibitem[{Krishna et~al.(2022)Krishna, Chang, Wieting, and Iyyer}]{Krishna2022RankGenIT}
Kalpesh Krishna, Yapei Chang, John Wieting, and Mohit Iyyer. 2022.
\newblock \href {https://api.semanticscholar.org/CorpusID:248887396} {Rankgen: Improving text generation with large ranking models}.
\newblock \emph{ArXiv}, abs/2205.09726.

\bibitem[{Liu et~al.(2023)Liu, Lin, Hewitt, Paranjape, Bevilacqua, Petroni, and Liang}]{Liu2023LostIT}
Nelson~F. Liu, Kevin Lin, John Hewitt, Ashwin Paranjape, Michele Bevilacqua, Fabio Petroni, and Percy Liang. 2023.
\newblock \href {https://api.semanticscholar.org/CorpusID:259360665} {Lost in the middle: How language models use long contexts}.
\newblock \emph{ArXiv}, abs/2307.03172.

\bibitem[{Liu et~al.(2021)Liu, Zheng, Demasi, Sabour, Li, Yu, Jiang, and Huang}]{Liu2021TowardsES}
Siyang Liu, Chujie Zheng, Orianna Demasi, Sahand Sabour, Yu~Li, Zhou Yu, Yong Jiang, and Minlie Huang. 2021.
\newblock \href {https://api.semanticscholar.org/CorpusID:235294326} {Towards emotional support dialog systems}.
\newblock \emph{ArXiv}, abs/2106.01144.

\bibitem[{Medeiros and Bosse(2018)}]{Medeiros2018UsingCF}
Lenin Medeiros and Tibor Bosse. 2018.
\newblock \href {https://api.semanticscholar.org/CorpusID:49397755} {Using crowdsourcing for the development of online emotional support agents}.
\newblock In \emph{Practical Applications of Agents and Multi-Agent Systems}.

\bibitem[{Peng et~al.(2022{\natexlab{a}})Peng, Hu, Xing, Xie, Sun, and Li}]{Peng2022ControlGU}
Wei Peng, Yue Hu, Luxi Xing, Yuqiang Xie, Yajing Sun, and Yunpeng Li. 2022{\natexlab{a}}.
\newblock \href {https://api.semanticscholar.org/CorpusID:248406141} {Control globally, understand locally: A global-to-local hierarchical graph network for emotional support conversation}.
\newblock In \emph{International Joint Conference on Artificial Intelligence}.

\bibitem[{Peng et~al.(2022{\natexlab{b}})Peng, Qin, Hu, Xie, and Li}]{Peng2022FADOFD}
Wei Peng, Ziyuan Qin, Yue Hu, Yuqiang Xie, and Yunpeng Li. 2022{\natexlab{b}}.
\newblock \href {https://api.semanticscholar.org/CorpusID:253244287} {Fado: Feedback-aware double controlling network for emotional support conversation}.
\newblock \emph{Knowl. Based Syst.}, 264:110340.

\bibitem[{Qin et~al.(2022)Qin, Feng, and Durme}]{Qin2022TheNT}
Guanghui Qin, Yukun Feng, and Benjamin~Van Durme. 2022.
\newblock \href {https://api.semanticscholar.org/CorpusID:246867127} {The nlp task effectiveness of long-range transformers}.
\newblock \emph{ArXiv}, abs/2202.07856.

\bibitem[{Reimers and Gurevych(2019)}]{Reimers2019SentenceBERTSE}
Nils Reimers and Iryna Gurevych. 2019.
\newblock \href {https://api.semanticscholar.org/CorpusID:201646309} {Sentence-bert: Sentence embeddings using siamese bert-networks}.
\newblock In \emph{Conference on Empirical Methods in Natural Language Processing}.

\bibitem[{Sharma et~al.(2021)Sharma, Lin, Miner, Atkins, and Althoff}]{Sharma2021TowardsFE}
Ashish Sharma, Inna~Wanyin Lin, Adam~S. Miner, David~C. Atkins, and Tim Althoff. 2021.
\newblock \href {https://api.semanticscholar.org/CorpusID:231639313} {Towards facilitating empathic conversations in online mental health support: A reinforcement learning approach}.
\newblock \emph{Proceedings of the Web Conference 2021}.

\bibitem[{Sharma et~al.(2020)Sharma, Miner, Atkins, and Althoff}]{Sharma2020ACA}
Ashish Sharma, Adam~S. Miner, David~C. Atkins, and Tim Althoff. 2020.
\newblock \href {https://api.semanticscholar.org/CorpusID:221761251} {A computational approach to understanding empathy expressed in text-based mental health support}.
\newblock \emph{ArXiv}, abs/2009.08441.

\bibitem[{Sun et~al.(2021)Sun, Krishna, Mattarella-Micke, and Iyyer}]{Sun2021DoLL}
Simeng Sun, Kalpesh Krishna, Andrew Mattarella-Micke, and Mohit Iyyer. 2021.
\newblock \href {https://api.semanticscholar.org/CorpusID:237572264} {Do long-range language models actually use long-range context?}
\newblock \emph{ArXiv}, abs/2109.09115.

\bibitem[{Touvron et~al.(2023)Touvron, Martin, Stone, Albert, Almahairi, Babaei, Bashlykov, Batra, Bhargava, Bhosale, Bikel, Blecher, Ferrer, Chen, Cucurull, Esiobu, Fernandes, Fu, Fu, Fuller, Gao, Goswami, Goyal, Hartshorn, Hosseini, Hou, Inan, Kardas, Kerkez, Khabsa, Kloumann, Korenev, Koura, Lachaux, Lavril, Lee, Liskovich, Lu, Mao, Martinet, Mihaylov, Mishra, Molybog, Nie, Poulton, Reizenstein, Rungta, Saladi, Schelten, Silva, Smith, Subramanian, Tan, Tang, Taylor, Williams, Kuan, Xu, Yan, Zarov, Zhang, Fan, Kambadur, Narang, Rodriguez, Stojnic, Edunov, and Scialom}]{Touvron2023Llama2O}
Hugo Touvron, Louis Martin, Kevin~R. Stone, Peter Albert, Amjad Almahairi, Yasmine Babaei, Nikolay Bashlykov, Soumya Batra, Prajjwal Bhargava, Shruti Bhosale, Daniel~M. Bikel, Lukas Blecher, Cristian~Cant{\'o}n Ferrer, Moya Chen, Guillem Cucurull, David Esiobu, Jude Fernandes, Jeremy Fu, Wenyin Fu, Brian Fuller, Cynthia Gao, Vedanuj Goswami, Naman Goyal, Anthony~S. Hartshorn, Saghar Hosseini, Rui Hou, Hakan Inan, Marcin Kardas, Viktor Kerkez, Madian Khabsa, Isabel~M. Kloumann, A.~V. Korenev, Punit~Singh Koura, Marie-Anne Lachaux, Thibaut Lavril, Jenya Lee, Diana Liskovich, Yinghai Lu, Yuning Mao, Xavier Martinet, Todor Mihaylov, Pushkar Mishra, Igor Molybog, Yixin Nie, Andrew Poulton, Jeremy Reizenstein, Rashi Rungta, Kalyan Saladi, Alan Schelten, Ruan Silva, Eric~Michael Smith, R.~Subramanian, Xia Tan, Binh Tang, Ross Taylor, Adina Williams, Jian~Xiang Kuan, Puxin Xu, Zhengxu Yan, Iliyan Zarov, Yuchen Zhang, Angela Fan, Melanie Kambadur, Sharan Narang, Aurelien Rodriguez, Robert Stojnic, Sergey Edunov, and
  Thomas Scialom. 2023.
\newblock \href {https://api.semanticscholar.org/CorpusID:259950998} {Llama 2: Open foundation and fine-tuned chat models}.
\newblock \emph{ArXiv}, abs/2307.09288.

\bibitem[{Tu et~al.(2022)Tu, Li, Cui, Wang, Wen, and Yan}]{Tu2022MISCAM}
Quan Tu, Yanran Li, Jianwei Cui, Bin Wang, Jiaxin Wen, and Rui Yan. 2022.
\newblock \href {https://api.semanticscholar.org/CorpusID:247748640} {Misc: A mixed strategy-aware model integrating comet for emotional support conversation}.
\newblock \emph{ArXiv}, abs/2203.13560.

\bibitem[{Turner et~al.(2023)Turner, Thiergart, Udell, Leech, Mini, and MacDiarmid}]{Turner2023ActivationAS}
Alexander~Matt Turner, Lisa Thiergart, David~S. Udell, Gavin Leech, Ulisse Mini, and Monte~Stuart MacDiarmid. 2023.
\newblock \href {https://api.semanticscholar.org/CorpusID:261049449} {Activation addition: Steering language models without optimization}.
\newblock \emph{ArXiv}, abs/2308.10248.

\bibitem[{Wu et~al.(2023)Wu, Yao, Chen, Pan, Wang, Liu, and Yu}]{Wu2023FromLM}
Xuansheng Wu, Wenlin Yao, Jianshu Chen, Xiaoman Pan, Xiaoyang Wang, Ninghao Liu, and Dong Yu. 2023.
\newblock \href {https://api.semanticscholar.org/CorpusID:263334329} {From language modeling to instruction following: Understanding the behavior shift in llms after instruction tuning}.
\newblock \emph{ArXiv}, abs/2310.00492.

\bibitem[{Zhao et~al.(2023)Zhao, Zhao, Wang, and Qin}]{Zhao2023TransESCSE}
Weixiang Zhao, Yanyan Zhao, Shilong Wang, and Bing Qin. 2023.
\newblock \href {https://api.semanticscholar.org/CorpusID:258547321} {Transesc: Smoothing emotional support conversation via turn-level state transition}.
\newblock In \emph{Annual Meeting of the Association for Computational Linguistics}.

\bibitem[{Zheng et~al.(2023{\natexlab{a}})Zheng, Chiang, Sheng, Zhuang, Wu, Zhuang, Lin, Li, Li, Xing, Zhang, Gonzalez, and Stoica}]{Zheng2023JudgingLW}
Lianmin Zheng, Wei-Lin Chiang, Ying Sheng, Siyuan Zhuang, Zhanghao Wu, Yonghao Zhuang, Zi~Lin, Zhuohan Li, Dacheng Li, Eric~P. Xing, Haotong Zhang, Joseph Gonzalez, and Ion Stoica. 2023{\natexlab{a}}.
\newblock \href {https://api.semanticscholar.org/CorpusID:259129398} {Judging llm-as-a-judge with mt-bench and chatbot arena}.
\newblock \emph{ArXiv}, abs/2306.05685.

\bibitem[{Zheng et~al.(2023{\natexlab{b}})Zheng, Liao, Deng, and Nie}]{Zheng2023BuildingES}
Zhonghua Zheng, Lizi Liao, Yang Deng, and Liqiang Nie. 2023{\natexlab{b}}.
\newblock \href {https://api.semanticscholar.org/CorpusID:261065100} {Building emotional support chatbots in the era of llms}.
\newblock \emph{ArXiv}, abs/2308.11584.

\end{thebibliography}

\appendix

\section{Datasets}
\subsection{ESConv Dataset Statistics}
\label{sec:appendix-ds-esconv}

Table \ref{tab:esconv-stat} summarizes some of the key statistics of the ESConv dataset.

\begin{table}[!ht]
    \centering
    \begin{tabular}{c|c}
        \hline
         \textbf{Category} & \textbf{Total}\\
         \hline
         \# dialogues & 1,053\\
         \# utterances & 31,410\\
         avg. length of dialogues & 29.8\\
         \# strategies & 8\\
         \hline
    \end{tabular}
    \caption{Some of the key statistics of the original ESConv dataset}
    \label{tab:esconv-stat}
\end{table}

\subsection{Extended ESConv Dataset Statistics}
\label{sec:appendix-ds-exesconv}

To create the extended version of the dataset for training our proposed models, we chose the most effective prompting template (c1\_hf) for strategy following and generated continuations by randomly selecting a strategy. To be more specific, we cut the conversations after "help seeker's" turn at some point between 5th and 23th turn of the conversation for 7b and 13b llama models and somewhere between 5th and 19th turn for 70b model. Afterwards, we randomly pick strategies with probability of 30\% and prompt the model to get the response. We then postprocess the responses by removing the indicators of the strategy or any unwanted textual span such as "Here is a response:". Table \ref{tab:exesc} summarizes the statistics of this dataset.

\begin{table*}
    \centering
    \begin{tabular}{c|c|c|c}
    \hline
         model name & number of conversations & number of continuations & min/max turn \\
         \hline
        llama-7b-chat  & 1,297 & 41,994 & 5/23 \\
        llama-13b-chat  & 1,297 & 41,822 & 5/23 \\
        llama-70b-chat & 1,297 & 24,760 & 5/19 \\
    \hline
    \end{tabular}
    \caption{statistics of the extended dialogue dataset}
    \label{tab:exesc}
\end{table*}

\subsubsection{Strategies and their definitions}

In tables \ref{tab:strategy-desc1} and \ref{tab:strategy-desc2} we provide all of the 15 strategies that we use to extend the dataset along with some examples of how they might be used. Both strategy and description will directly be used inside of the prompt.

% \subsubsection{Statistics}

% As mentioned earlier we continue the conversations by randomly choosing a few strategies from the pool of strategies. More specifically, for each strategy we will continue the conversation using that strategy with the probability of 30\%.

\begin{figure*}[!ht]
    \centering
    \includegraphics[width=0.95\textwidth]{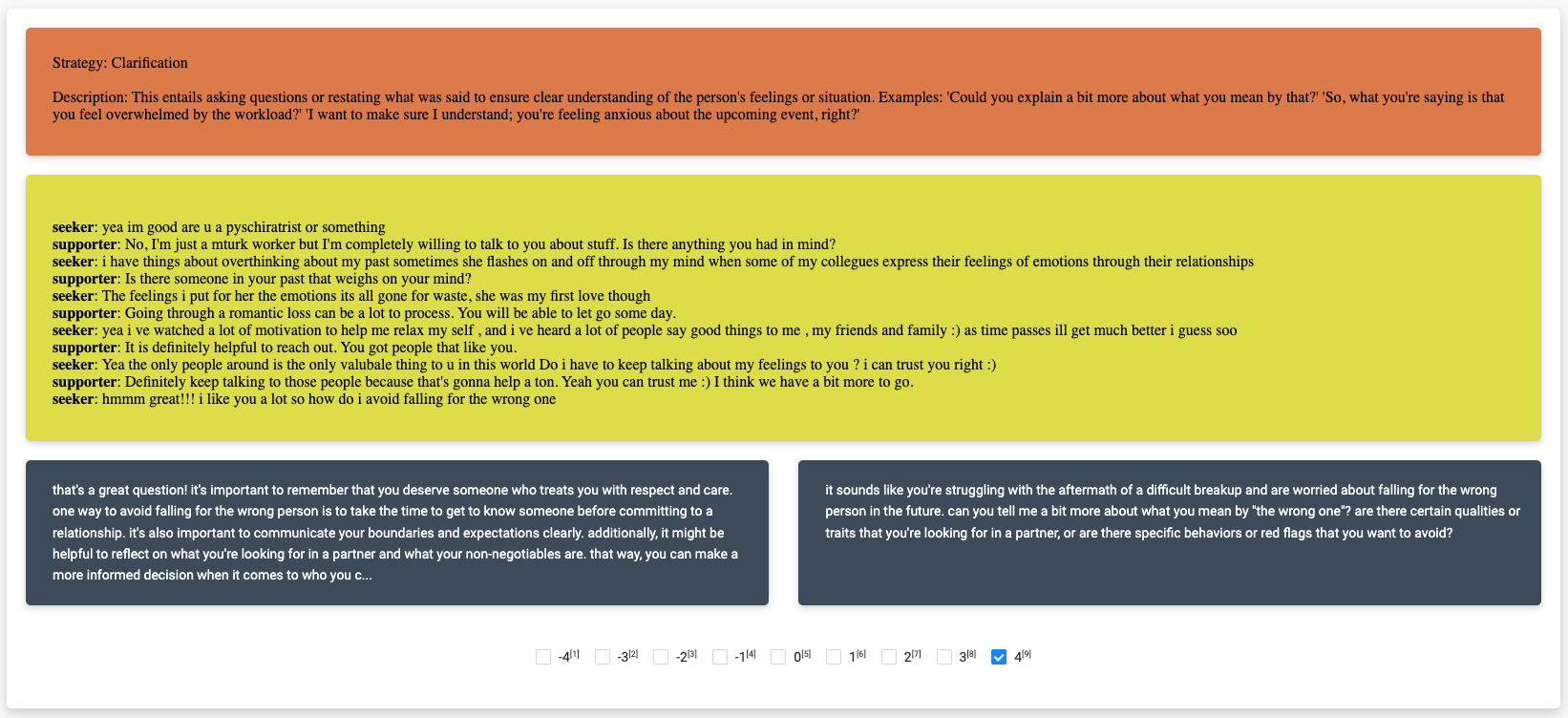} % Adjust the path to your image file and the scaling as needed
    \caption{A sample annotation task}
    \label{fig:annotation} % Label for referencing the figure in text
\end{figure*}

% \begin{figure}[!ht]
%     \centering
%     \includegraphics[width=0.47\textwidth]{figs/modelsizeSRA.png} % Adjust the path to your image file and the scaling as needed
%     \caption{Analyzing SRA given different prompt templates indicates that the position of the strategy guidelines inside of the prompt significantly influences the amount of attention that the model pays to the strategy tokens. It can be seen that the c1\_hf template receives the most SRA regardless of the model size}
%     \label{fig:modelsizeSRA} % Label for referencing the figure in text
% \end{figure}

% \section{Consistency of SRA Across Different Strategies}

% We also provide an analysis of the SRA metric across different strategies using llama2-70b-chat model and all the 7 prompting baselines. We observe the same pattern as the aggregated SRA shown in figure \ref{fig:modelsizeSRA} for each of the strategies. For this analysis we used the same collections described in section \ref{sec:sampling}. Figure \ref{fig:perstrategySRA} depicts the results of this analysis.

% \begin{figure*}[!ht]
%     \centering
%     \includegraphics[width=0.95\textwidth]{figs/perstrategySRA.png} % Adjust the path to your image file and the scaling as needed
%     \caption{Per strategy SRA for different prompt template responses for the llama-70b-chat model}
%     \label{fig:perstrategySRA} % Label for referencing the figure in text
% \end{figure*}

\section{Annotation Task Details}
\label{appendix-annotation}
Figure \ref{fig:annotation} shows a sample annotation task. Two of the authors of the paper perform the annotation task. We simply instruct the annotators with the following paragraph before starting the annotation: 

\texttt{On the top of each task you will see a strategy along with it's definition. Afterwards you will be given a conversation between an emotional supporter (counselor) and a person who is seeking help. The conversation is cut at a random spot with help seeker uttering the last turn. Then you will see two continuations of the conversation using the proposed strategy. Your task is to choose the continuation that best follows the strategy while maintaining consistency and coherence with respect to the conversation history and remains natural and conversational.
You have 8 options for scoring +4 meaning the right continuation is extremely preferred over the left continuation and vice versa. If none of the responses satisfy the requirement or both of them are perfectly following the strategy and are consistent and coherent, choose 0 but if one of them is slightly better lean your score towards that answer accordingly.}

\begin{figure*}
    \centering
    \includegraphics[width=1\linewidth]{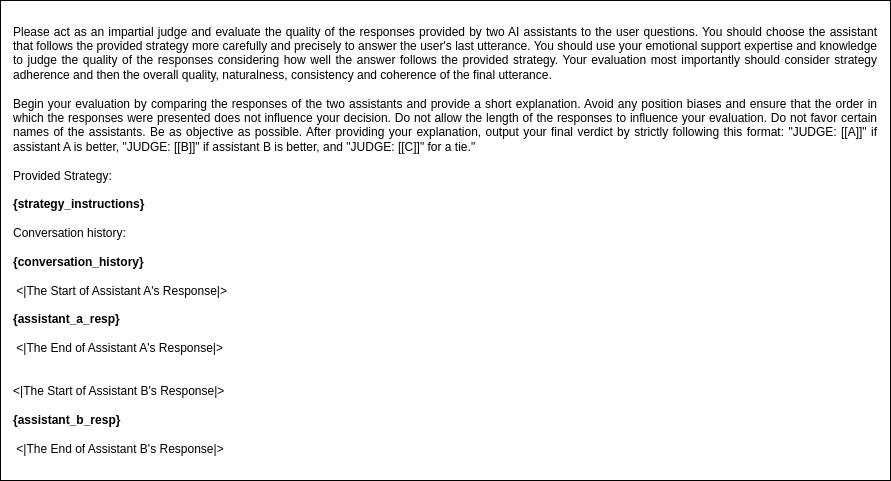}
    \caption{The prompt template used for comparison of two model responses using gpt-4o}
    \label{fig:eval-prompt}
\end{figure*}

\section{Model Based Evaluation}
\label{appendix:model-based-eval}

We use the prompt template shown in figure \ref{fig:eval-prompt}. More specifically, we ask the model to act as an impartial judge and compare the strategy adherence of the two models while taking into account the consistency, coherence, quality and the response being natural. The general judging template is adopted from MT-Bench \cite{Zheng2023JudgingLW} paper. 

We incorporate conversation histories of two different models, the intended strategy instructions and the responses by both models into the judge prompt. We also alternate between the position of the both models' responses in the judge prompt and call the judge twice to make sure we avoid positional bias. The judge will then have to first reason about which model is potentially better and then choose either one of them or call it a tie.

\begin{figure*}
    \centering
    \includegraphics[width=\textwidth]{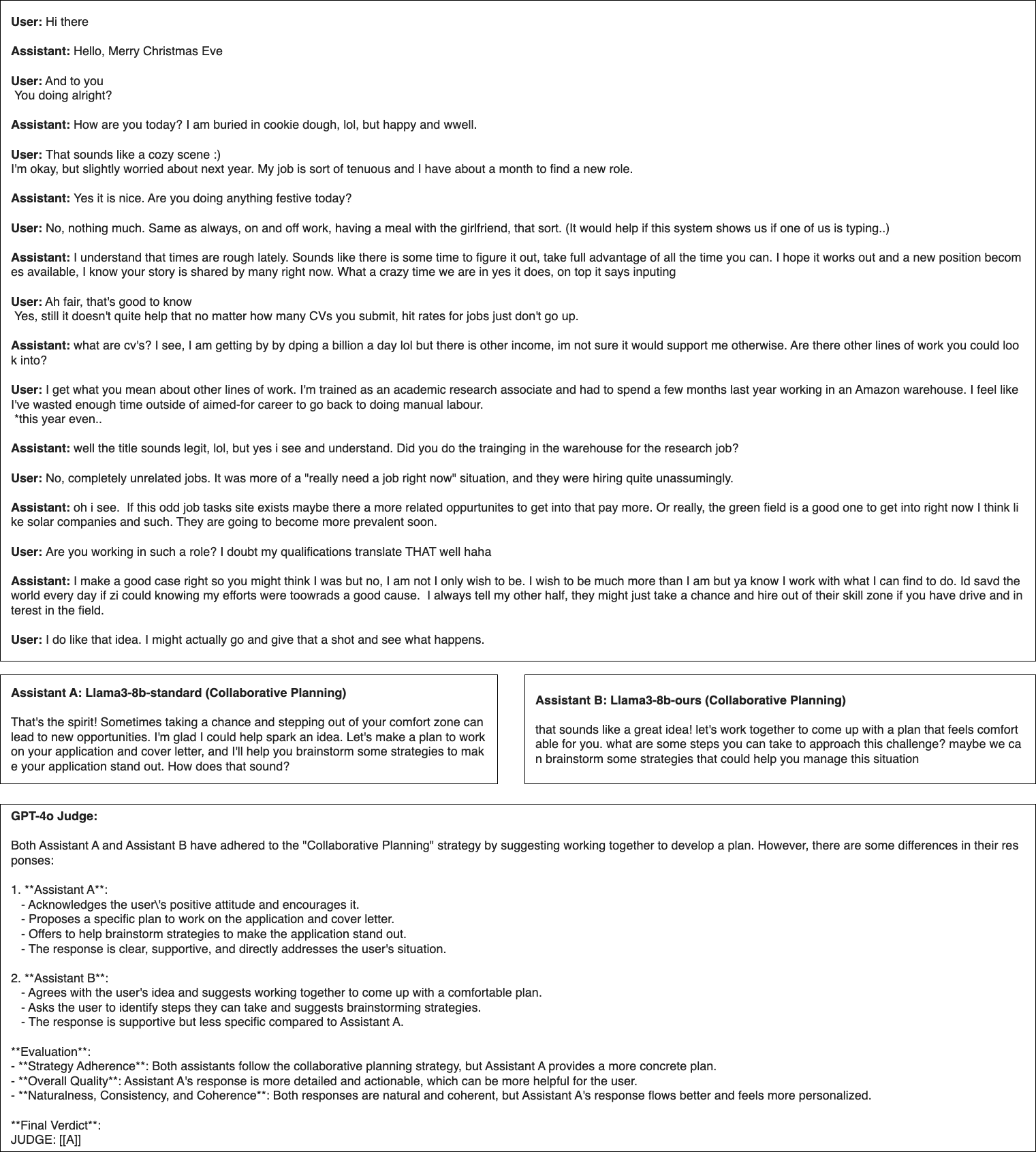} % Adjust the path to your image file and the scaling as needed
    \caption{An example where gpt-4o judges the base model to give a better response that both aligns with the strategy and is more natural, coherent and consistent}
    \label{fig:lose-ex} % Label for referencing the figure in text
\end{figure*}

\begin{figure*}
    \centering
    \includegraphics[width=\textwidth]{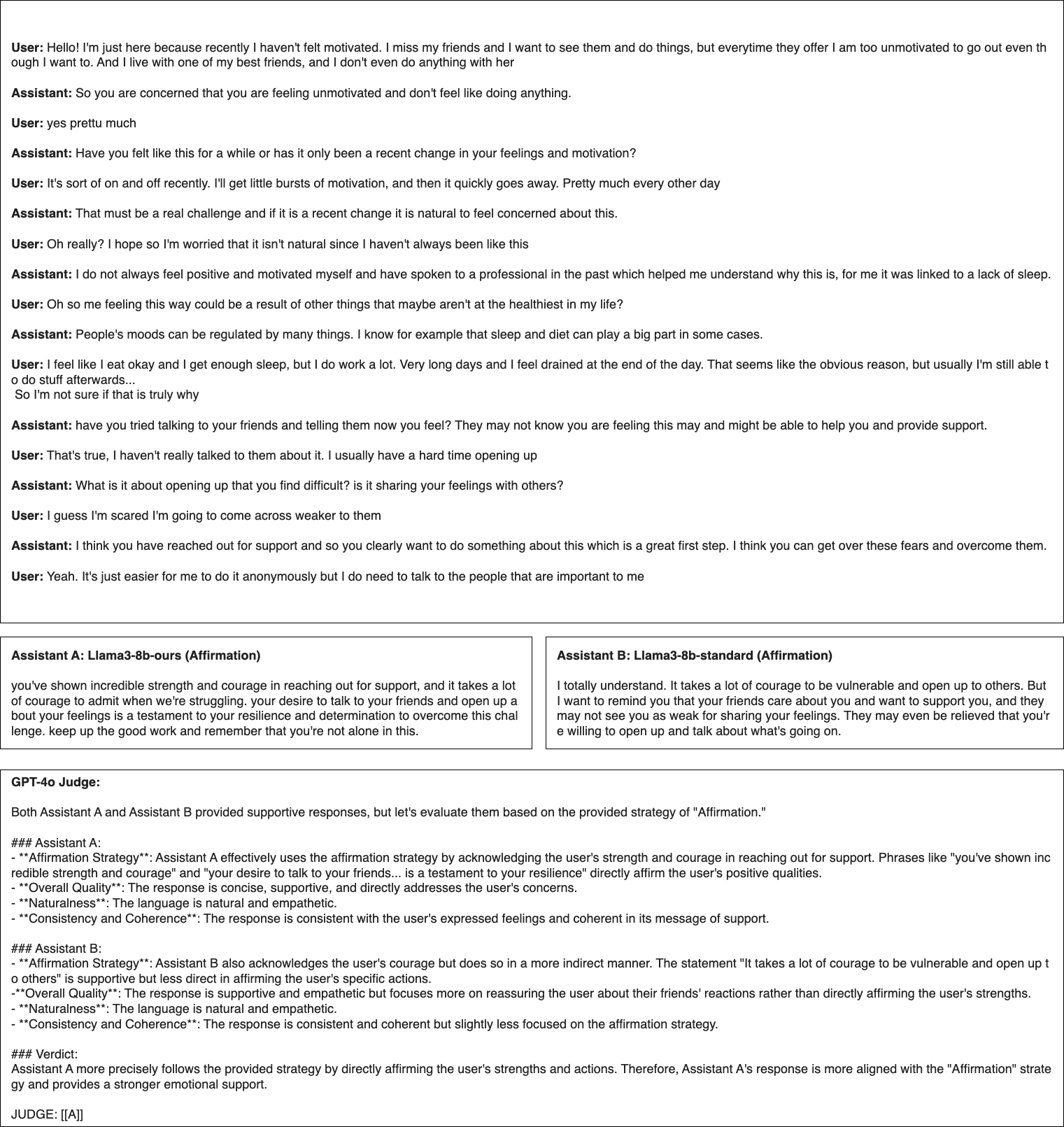} % Adjust the path to your image file and the scaling as needed
    \caption{An example where gpt-4o judges the our model to give a better response that both aligns with the strategy and is more natural, coherent and consistent}
    \label{fig:win-ex} % Label for referencing the figure in text
\end{figure*}

Figure \ref{fig:win-ex} illustrates an examples where our proposed llama3-8b-instruct model outperforms the base model and figure \ref{fig:lose-ex} shows an example where our model loses to the base model because of lack of natural flow in the response.

% \section{Prompt Construction}
% \label{appendix:prompt}
% \begin{figure}
%     \centering
%     \includegraphics[width=0.47\textwidth]{figs/standardprompt.png} % Adjust the path to your image file and the scaling as needed
%     \caption{The formation of the standard prompting baseline.}
%     \label{fig:standardprompt} % Label for referencing the figure in text
% \end{figure}

\section{Predictability of the Strategy from the Response}
\label{sec:appendix_pred}

This section explores the assumption that the effectiveness of a model in following a given strategy can be quantified by assessing how predictable the strategy is, given the generated utterance. We hypothesize that there is a direct correlation between the predictability of the strategy and the model's adherence to it. \textbf{Although predictability of the responses does not necessarily indicate the adherence to the specific strategy, it perfectly assess the ability of different methods in distinguishing between strategies when generating the response.}

To formalize this concept, we utilize Bayes' rule, a fundamental theorem in probability theory. Bayes' rule describes the probability of an event based on prior knowledge of conditions related to the event. In our context, it is used to relate the probability of a strategy \( S \) given a generated response \( R \), to the probability of generating a response given a strategy. The rule is formulated as:

\begin{equation}
    P(S | R) = \frac{P(R | S) \times P(S)}{P(R)}
\end{equation}

Here, \( P(S | R) \) represents the posterior probability, indicating the likelihood of the strategy \( S \) given the observation of the response \( R \). \( P(R | S) \) is the likelihood of generating the response \( R \) when following the strategy \( S \). \( P(S) \) and \( P(R) \) are the prior probabilities of the strategy and the response, respectively.

A high posterior probability, \( P(S | R) \), suggests that the response \( R \) strongly indicates the use of strategy \( S \), implying effective adherence by the model to the strategy. Conversely, a low value indicates weaker adherence to the strategy.

\subsection{Measuring predictability based on lexical features}
Our first proposal is a baseline model using Bag of Words Logistic Regression over N-grams to identify lexical differences between different templates' responses. This model is selected for its simplicity and interpretability. It allows us to easily understand which words or phrases significantly contribute to the distinctiveness of the responses. The model is defined as:

\begin{equation}
    P(S|R) = \sigma\left(\sum_{i=1}^{N} \omega_i \cdot x_i + b\right)
\end{equation}

where \( \sigma \) is the sigmoid function, \( \omega_i \) are the weights assigned to each n-gram, \( x_i \) are the n-gram features extracted from the response, and \( b \) is the bias term. We remove English stop words and words that appear in more than 90\% of the responses and then build 2-gram and 3-gram feature vectors to train the logistic regression model.

\subsection{Measuring predictability based on semantic features}
To complement the first model and capture deeper semantic features, we also employ a Sentence Bert model \cite{Reimers2019SentenceBERTSE} for sequence classification. To be specific, we use \emph{all-mpnet-base-v2} model which stands on top of the leader board for the best quality of sentence encodings over 14 tasks in different domains \footnote{\href{https://www.sbert.net/docs/pretrained\_models.html}{https://www.sbert.net/docs/pretrained\_models.html}}. This model provides us with the capability to discern intricate semantic patterns that might be overlooked by the simpler lexical predictor. We first employ the Sentence Bert model according to equation \ref{eq:sbert} where $R$ is the sequence of response tokens and retrieve an aggregate embedding for the whole response (in case of mpnet model we use, it will be a normalized average of the embeddings of all tokens in the sequence). Afterwards, same as what we did with the lexical predictor, we feed the encoding to a logistic regression model to predict the strategy class.

\begin{align}
\label{eq:sbert}
    X &= \text{Normalize}\left(\text{Mean}(\text{SBERT}(R))\right), \\
    P(S|R) &= \sigma\left(\sum_{i=1}^{N} \omega_i \cdot x_i + b\right),
\end{align}

\begin{figure}
    \centering
    \includegraphics[width=0.47\textwidth]{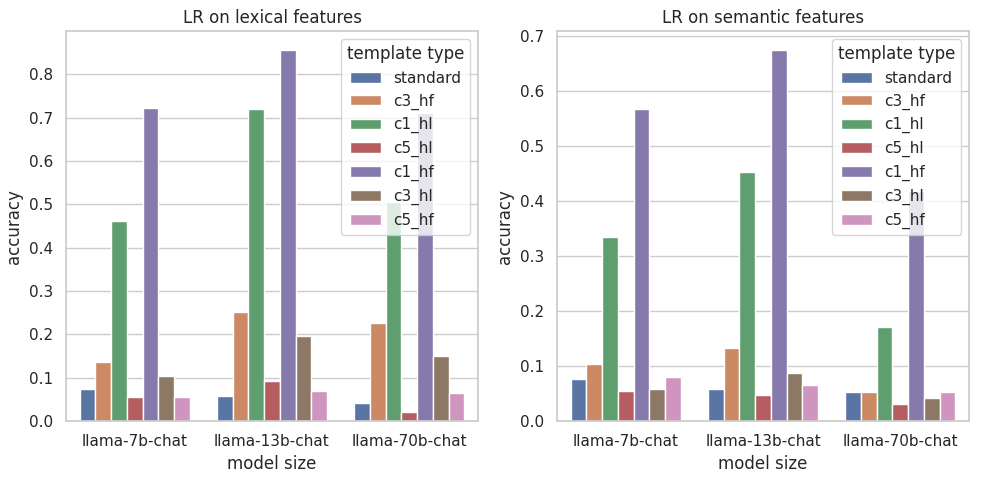} % Adjust the path to your image file and the scaling as needed
    \caption{Comparison of the predictability of the strategy of different prompt responses across different model sizes. We report accuracy of prediction using two predictors one operating on lexical features of response and the other one on semantic features of the response.}
    \label{fig:pred-eval} % Label for referencing the figure in text
\end{figure}

\subsection{Predictability results}
We measure the predictability of response strategy in the responses provided by all of the baselines on the held out test set using llama-2 suit of models. We randomly split each collection to 80/20 portions of training and test and train both mentioned models using 4-fold cross validation and report the prediction accuracies on the test set. We observe that the predictability of the responses in one collection is highly correlated with SRA of the responses in that collection. 

Figure \ref{fig:pred-eval} show the accuracy of the predictors trained on each of the extended data collections corresponding to different baseline models using bag of word embeddings and sentence bert embeddings of responses and a logistic regression classifier. 

By qualitatively analyzing the coefficients of the logistic regression model trained on lexical features, we observe that not only the responses given by the high SRA prompts are predictable (distinguishable) but also the high coefficient n-grams are completely relevant to the class of the strategy. We show a qualitative analysis of the lexical predictor trained on the responses of the llama2-13b model using \emph{c1\_hf} prompt template. After training the logistic regression model on training portion of the responses using bag-of-words features, we report top-5 features with highest coefficient in table \ref{tab:top5_ngrams}. According to this analysis, not only the responses are distinguishable, but also highest coefficients are corresponding to relevant phrases that can explain the strategy class. For instance, in the \textbf{Collaborative Planning} class, top coefficients contain phrases such as "work together" and "brainstorm strategies".

\section{Classifier Error Analysis}
\label{appendix:classifer}

\begin{table}[htbp]
\centering
\resizebox{\columnwidth}{!}{
\begin{tabular}{|l|c|c|c|c|}
\hline
\textbf{Strategy} & \textbf{Precision} & \textbf{Recall} & \textbf{F1-Score} & \textbf{Support} \\ \hline
Affirmation & 0.99 & 1.00 & 0.99 & 74 \\ \hline
Avoid Judgment and Criticism & 0.75 & 0.92 & 0.83 & 62 \\ \hline
Chit Chat & 0.91 & 1.00 & 0.95 & 63 \\ \hline
Clarification & 0.98 & 0.82 & 0.90 & 57 \\ \hline
Collaborative Planning & 0.96 & 0.99 & 0.97 & 75 \\ \hline
Emotional Validation & 0.98 & 0.97 & 0.98 & 65 \\ \hline
Normalize Experiences & 0.91 & 0.98 & 0.95 & 54 \\ \hline
Offer Hope & 0.97 & 0.94 & 0.95 & 63 \\ \hline
Promote Self-Care Practices & 0.97 & 1.00 & 0.99 & 72 \\ \hline
Provide Different Perspectives & 0.96 & 0.96 & 0.96 & 67 \\ \hline
Reflective Statements & 0.90 & 0.91 & 0.91 & 81 \\ \hline
Reframe Negative Thoughts & 0.99 & 0.92 & 0.95 & 76 \\ \hline
Share Information & 0.95 & 0.91 & 0.93 & 67 \\ \hline
Stress Management & 0.98 & 0.88 & 0.93 & 60 \\ \hline
Suggest Options & 0.93 & 0.88 & 0.90 & 64 \\ \hline
\end{tabular}
}
\caption{Precision, Recall, F1-Score, and Support for Each Strategy}
\label{tab:clsprf}
\end{table}

For evaluating the performance of the classifier, we use the held out 100 conversations. We randomly break conversations the same way we outlined in section \ref{sec:exesc} and randomly pick our fine-tuned llama2-7b-chat and llama3-8b-instruct models to continue the conversation using a random strategy. Then we collect 1000 responses. Afterwards, we apply our classifier on the responses and measure the inferred strategy given the response. 
The authors of the paper are asked to verify if the inferred class of strategy is used in the response generated by the model. The exact annotation instruction is as follows:

\emph{Given the response bellow, verify if the response uses the mentioned strategy. If the response uses multiple strategies, only verify it's correctness if the strategy forms the main argument in the response.}

the classifier achieves 93.6\% accuracy on this task. Table \ref{tab:clsprf} shows the detailed performance on different classes. It can be observed that the classifier mostly struggles in \emph{Avoid Judgment and Criticism} class.

\begin{table*}
    \centering
    \begin{tabularx}{\textwidth}{|p{3cm}|X|}
    \hline
        \textbf{strategy} & \textbf{description}\\
        \hline
         Affirmation & This involves acknowledging and positively reinforcing an individual's strengths, feelings, or actions. Examples: 'You've shown incredible resilience in facing these challenges.' 'I admire your dedication to improving your situation.' 'Your ability to stay hopeful in tough times is truly commendable.'\\
         \hline
         Clarification & This entails asking questions or restating what was said to ensure clear understanding of the person's feelings or situation. Examples: 'Could you explain a bit more about what you mean by that?' 'So, what you're saying is that you feel overwhelmed by the workload?' 'I want to make sure I understand; you're feeling anxious about the upcoming event, right?'\\
         \hline
         Collaborative Planning & This involves working together to develop strategies or plans to address specific issues or challenges. Examples: 'Let's brainstorm some strategies that could help you manage this stress.' 'We can work together to come up with a plan that feels comfortable for you.' 'How about we outline some steps you can take to approach this problem?'\\
         \hline
         Emotional Validation & This strategy involves acknowledging and accepting the person's emotions as legitimate and important. Examples: 'It's completely normal to feel sad in a situation like this.' 'Your feelings of frustration in this case are absolutely understandable.' 'I hear you, and it makes sense that you would feel anxious about this.'\\
         \hline
         Normalize Experiences & This approach helps the person understand that their experiences or feelings are common and not something to be ashamed of. Examples: 'Many people go through similar challenges, and it's okay to feel this way.' 'Feeling overwhelmed in such situations is a common reaction.' 'It's normal to have ups and downs in response to life's stresses.'\\
         \hline
         Offer Hope& This involves providing reassurance that things can improve and that there is hope for a better future. Examples: 'I'm confident that you'll find a way through this challenge.' 'Things might be tough now, but there is always a possibility for change and growth.' 'I believe in your ability to overcome these obstacles.'\\ 
         \hline
         Promote Self-Care Practices& Encouraging the person to engage in activities that promote physical, emotional, and mental well-being. Examples: 'Have you considered setting aside some time for relaxation or a hobby you enjoy?' 'Taking care of your health is important, maybe try some exercise or meditation.' 'Remember to take breaks and do things that make you feel good.'\\
         \hline
         Provide Different Perspectives & Offering new viewpoints or ways of thinking about a situation to help broaden understanding and possibly reduce distress. Examples: 'Have you considered looking at the situation from this angle?' 'Sometimes, stepping back and viewing things differently can be helpful.' 'What if we think about the potential positive outcomes of this scenario?'\\
         \hline
        
    \end{tabularx}
        \caption{Strategy 1 to 8 along with their descriptions}
    \label{tab:strategy-desc1}
\end{table*}

\begin{table*}
    \centering
    \begin{tabularx}{\textwidth}{|p{3cm}|X|}
    \hline
        \textbf{strategy} & \textbf{description}\\
        \hline
         
        Avoid Judgment and Criticism & This strategy focuses on providing support without expressing negative judgments or criticisms of the person's thoughts, feelings, or actions. Examples: 'It's understandable that you felt that way in that situation.' 'Everyone makes mistakes, and it's okay to be imperfect.' 'Your feelings are valid, and it's okay to express them.'\\
         \hline
         Reflective Statements& Mirroring back what the person has said to show understanding and empathy. Examples: 'It sounds like you're feeling really overwhelmed by your workload.' 'You seem to be saying that this situation has made you feel anxious.' 'I hear that you're finding it hard to cope with these changes.'\\
         \hline
         Reframe Negative Thoughts& Helping to shift negative or unhelpful thought patterns into more positive or realistic ones. Examples: 'Instead of thinking of it as a failure, could we see it as a learning opportunity?' 'What if we try to focus on what you can control in this situation?' 'Let's look for the strengths you've shown in dealing with this.'\\
         \hline
         Share Information& Providing factual information or resources that might be helpful in understanding or coping with a situation. Examples: 'I read an article about coping strategies that might be useful for you.' 'There are some great books that offer insights into managing these feelings.' 'I can share some websites that provide helpful tips on stress management.'\\
         \hline
         Stress Management& Offering techniques or suggestions to help reduce or manage stress. Examples: 'Have you tried deep breathing or mindfulness exercises to manage stress?' 'Creating a regular routine can sometimes help in reducing stress levels.' 'Exercise can be a great way to relieve stress and improve mood.'\\
         \hline
         Suggest Options& Presenting various possibilities or alternatives that the person might consider in their situation. Examples: 'One option might be to talk to someone you trust about what you're going through.' 'Have you thought about joining a support group for this issue?' 'Maybe trying a new approach to this problem could yield different results.'\\
         \hline
         Chit Chat& Engaging in light, casual conversation to build rapport and provide a sense of normalcy and comfort. Examples: 'How's your day going so far?' 'Did you see that funny movie that came out recently?' 'I love this weather we're having. Do you enjoy outdoor activities?'\\
         \hline
    \end{tabularx}
    \caption{Strategy 9 to 15 along with their descriptions}
    \label{tab:strategy-desc2}
\end{table*}

\begin{table*}
\centering
\begin{tabular}{|l|p{7cm}|}
\hline
\textbf{Strategy}                      & \textbf{Top 5 N-grams}                                         \\ \hline
Affirmation                            & truly commendable, takes lot, shown incredible, strength resilience, resilience facing   \\ \hline
Avoid Judgment and Criticism           & important remember, okay feel, remember everyone, completely understandable, understandable feeling  \\ \hline
Chit Chat                              & day going, oh gosh, outdoor activities, oh goodness, hey day    \\ \hline
Clarification                          & tell mean, clarify saying, tell bit, clarify feeling, feeling overwhelmed     \\ \hline
Collaborative Planning                 & work together, together come, let work, come plan, brainstorm strategies  \\ \hline
Emotional Validation                   & completely understandable, valid important, normal feel, completely normal, absolutely valid \\ \hline
Normalize Experiences                  & many people, okay feel, completely normal, important remember, normal feel        \\ \hline
Offer Hope                             & better future, hope better, want know, believe ability, find way    \\ \hline
Promote Self-Care Practices            & aside time, setting aside time, considered setting, hobby enjoy, time relaxation \\ \hline
Provide Different Perspectives         & instead focusing, different perspective, considered looking, situation different, additionally might \\ \hline
Reflective Statements                  & sounds like, like feeling, understandable feeling, feeling really, tell feeling     \\ \hline
Reframe Negative Thoughts              & instead focusing, try reframe, let try, reframe opportunity, let focus \\ \hline
Share Information                      & resources available, additionally many, online resources, many resources, might helpful \\ \hline
Stress Management                      & deep breathing, manage stress, regular routine, techniques help, stress levels \\ \hline
Suggest Options                        & option could, one option, additionally might, another option, option might     \\ \hline
\end{tabular}
\caption{Top 5 3-gram and 2-gram features for strategy classification in lexical predictor}
\label{tab:top5_ngrams}
\end{table*}

\section{Steerability analysis of Llama-3 models}
\label{appendix:llama3}
\begin{figure*}[ht]
    \centering
    \includegraphics[width=\textwidth]{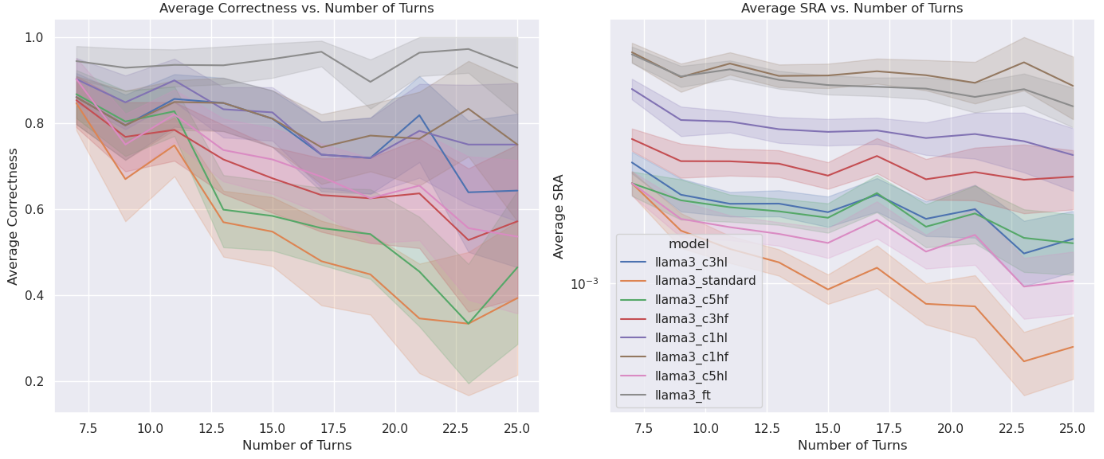} % Adjust the path to your image file and the scaling as needed
    \caption{Left: average accuracy of the strategy following for each model with respect to the turn of the conversation, Right: average SRA of the responses with respect to the turn of the conversation}
    \label{fig:depth2} % Label for referencing the figure in text
\end{figure*}

We fix the same experimental setup as in section \ref{sec:eval_long} for llama3-8b-instruct and measure the average accuracy and SRA of the model responses in depth of the conversation. Although llama-3 models go through a more extensive phase of steerability training \citep{dubey2024llama3herdmodels}, we still observe the lack of adherence to the prompted strategy for these models and therefore a huge gap between the performance of the base llama3-8b-instruct model and our fine-tuned counter part in following the strategy. Figure \ref{fig:depth2} demonstrates the accuracy and SRA of the responses by different llama3-8b based models in depth of the conversation.

\end{document}